\documentclass{article}

    \PassOptionsToPackage{numbers, compress}{natbib}
\usepackage[final]{neurips_2025}

\usepackage[utf8]{inputenc} % allow utf-8 input
\usepackage[T1]{fontenc}    % use 8-bit T1 fonts
\usepackage{url}            % simple URL typesetting
\usepackage{booktabs}       % professional-quality tables
\usepackage{amsfonts}       % blackboard math symbols
\usepackage{nicefrac}       % compact symbols for 1/2, etc.
\usepackage{microtype}      % microtypography
\usepackage[format=plain,labelformat=simple,labelsep=period,font=small,compatibility=false]{caption}
\usepackage[font=footnotesize,skip=3pt,subrefformat=parens]{subcaption}
\usepackage{graphicx}
\usepackage{amsmath}
\usepackage{amssymb}
\usepackage[bb=dsserif]{mathalpha}  % to have $\mathbb 1$

\usepackage{afterpage}  % \afterpage
\usepackage{multirow}
\usepackage{makecell}
\usepackage{placeins}  % FloatBarrier
\usepackage{ifthen}  % provides conditonals...
\usepackage{diagbox}  % to have diagonal line in table cell
\usepackage[hang,flushmargin,bottom]{footmisc}  % footnote noindent
\usepackage[table,xcdraw]{xcolor}
\usepackage[capbesideposition=outside,capbesidesep=quad]{floatrow}
\usepackage[export]{adjustbox}

\usepackage[american]{babel}  % required by csquotes
\usepackage[autostyle=true, english=american]{csquotes}
\MakeOuterQuote{"}  % to automatically handle `` ''  =>  just use " / '

\usepackage{algorithm}
\usepackage{listings}
\definecolor{citecolor}{HTML}{3270b5}
\usepackage[pagebackref,breaklinks,colorlinks,citecolor=citecolor]{hyperref}
\usepackage[capitalize]{cleveref}
\crefname{section}{Sec.}{Secs.}
\Crefname{section}{Section}{Sections}
\Crefname{table}{Table}{Tables}
\crefname{table}{Tab.}{Tabs.}
\crefname{algorithm}{Alg.}{Algs.}

\definecolor{red}{HTML}{cc1100}
\definecolor{green}{HTML}{39b54a}  % green
\definecolor{blue}{HTML}{004bb3}

\definecolor{darkred}{HTML}{ea4335}
\definecolor{darkgreen}{HTML}{228B22}
\definecolor{darkblue}{HTML}{35394B}

\definecolor{orange}{HTML}{cc7700}
\definecolor{gray}{HTML}{efefef}
\definecolor{darkgray}{HTML}{808080}
\definecolor{lightpurple}{HTML}{a56dba}

\definecolor{BestBlue}{HTML}{1F77B4}   % elegant blue (Matplotlib “tab:blue”)
\newcommand{\best}[1]{\cellcolor{BestBlue!15}\textbf{#1}}   % best (slightly deeper tint + bold)
\newcommand{\second}[1]{\cellcolor{BestBlue!9}#1}           % second-best (lighter tint)

\newcommand{\green}[1]{{\color{green}#1}}
\newcommand{\darkgray}[1]{{\color{darkgray}#1}}

\definecolor{scratch}{HTML}{001219}

\definecolor{pretrain}{HTML}{0a9396}

\definecolor{deemph}{gray}{0.6}

\definecolor{baselinecolor}{gray}{.9}
\newcommand{\baseline}[1]{\cellcolor{baselinecolor}{#1}}

\newcommand{\plus}{\textcolor{darkblue}{\raisebox{0.3ex}{\tiny +\,}}}

\newcommand{\up}[2][]{%
  \ifthenelse { \equal {#1} {} }
  {\makebox[-2pt][l]{~\green{\fontsize{7pt}{1em}\selectfont$\uparrow$#2}}}
  {\makebox[-2pt][l]{\csname#1\endcsname{ \fontsize{7pt}{1em}\selectfont$\uparrow$#2}}}
}
\newcommand{\down}[2][]{%
  \ifthenelse { \equal {#1} {} }
  {\makebox[-2pt][l]{~{\fontsize{7pt}{1em}\selectfont$\downarrow$#2}}}
  {\makebox[-2pt][l]{\csname#1\endcsname{ \fontsize{7pt}{1em}\selectfont$\downarrow$#2}}}
}

\newcolumntype{x}[1]{>{\centering\arraybackslash}p{#1}}
\newcolumntype{y}[1]{>{\raggedright\arraybackslash}p{#1}}
\newcolumntype{z}[1]{>{\raggedleft\arraybackslash}p{#1}}
\newlength\savewidth
\newcommand{\tablestyle}[2]{\setlength{\tabcolsep}{#1}\renewcommand{\arraystretch}{#2}\centering\footnotesize}
\renewcommand{\paragraph}[1]{\vspace{0mm}\noindent\textbf{#1}}

\newcommand{\app}{\raise.17ex\hbox{$\scriptstyle\sim$}}

\makeatletter
\DeclareRobustCommand\onedot{\futurelet\@let@token\@onedot}
\def\@onedot{\ifx\@let@token.\else.\null\fi\xspace}
\def\eg{\emph{e.g}\onedot}

\usepackage{bm}

\def\1{\bm{1}}

\newcommand*{\@rowstyle}{}
\newcommand*{\rowstyle}[1]{% sets the style of the next row
  \gdef\@rowstyle{#1}%
  \@rowstyle\ignorespaces%
}
\newcolumntype{=}{% resets the row style
  >{\gdef\@rowstyle{}}%
}
\newcolumntype{+}{% adds the current row style to the next column
  >{\@rowstyle}%
}

\usepackage{xspace}
\newcommand{\method}{GEM\xspace}

\newcommand{\starfootnotetext}[1]{%
  \begingroup
    \renewcommand\thefootnote{\fnsymbol{footnote}}%
    \footnotetext[1]{#1}% 1 renders as *
  \endgroup
}

\title{
On Geometry-Enhanced Parameter-Efficient Fine-Tuning for 3D Scene Segmentation
}

\author{%
Liyao Tang \\
The University of Sydney \\
\texttt{\small ltan9687@uni.sydney.edu.au} \\
\And
Zhe Chen$^{1*}$ \\
La Trobe University \\
\texttt{\small zhe.chen@latrobe.edu.au} \\
\And
Dacheng Tao$^{2*}$ \\
\hspace{-.93em}Nanyang Technological University \\
\texttt{\small dacheng.tao@gmail.com} \\
\vspace{-10pt}
}

\AtBeginDocument{\colorlet{defaultcolor}{.}}  % capture default text color

\begin{document}
\let\svthefootnote\thefootnote  % save footnote counter

% - check default setting:
% \the\baselineskip     10.95pt
% \the\textfloatsep     20.0pt plus 2.0pt minus 4.0pt
% \the\floatsep         12.0pt plus 2.0pt minus 2.0pt
% \the\intextsep        12.0pt plus 2.0pt minus 2.0pt
% \the\dbltextfloatsep  20.0pt plus 2.0pt minus 4.0pt

% \the\textfloatsep     20.0pt plus 2.0pt minus 4.0pt
% \the\dbltextfloatsep  20.0pt plus 2.0pt minus 4.0pt
% \the\floatsep         12.0pt plus 2.0pt minus 2.0pt

% - set tighter length:
\setlength{\textfloatsep}{10.0pt plus 0.0pt minus 4.0pt}
\setlength{\dbltextfloatsep}{8.0pt plus 2.0pt minus 4.0pt}
\setlength{\floatsep}{8.0pt plus 0.0pt minus 4.0pt}

\maketitle
\starfootnotetext{Corresponding authors.}
\footnotetext[1]{Also affiliated with Cisco - La Trobe Centre for AI and IoT.}
\footnotetext[2]{College of Computing and Data Science, NTU, Singapore 639798.}
\setcounter{footnote}{0}

\begin{abstract}
The emergence of large-scale pre-trained point cloud models has significantly advanced 3D scene understanding, but adapting these models to specific downstream tasks typically demands full fine-tuning, incurring high computational and storage costs. Parameter-efficient fine-tuning (PEFT) techniques, successful in natural language processing and 2D vision tasks, would underperform when naively applied to 3D point cloud models due to significant geometric and spatial distribution shifts.
Existing PEFT methods commonly treat points as orderless tokens, neglecting important local spatial structures and global geometric contexts in 3D modeling.
To bridge this gap, we introduce the Geometric Encoding Mixer (GEM), a novel geometry-aware PEFT module specifically designed for 3D point cloud transformers. GEM explicitly integrates fine-grained local positional encodings with a lightweight latent attention mechanism to capture comprehensive global context, thereby effectively addressing the spatial and geometric distribution mismatch.
Extensive experiments demonstrate that GEM achieves performance comparable to or sometimes even exceeding full fine-tuning, while only updating 1.6\% of the model's parameters, fewer than other PEFT methods.
With significantly reduced training time and memory requirements, our approach thus sets a new benchmark for efficient, scalable, and geometry-aware fine-tuning of large-scale 3D point cloud models.
Code is available at \href{https://github.com/LiyaoTang/GEM}{https://github.com/LiyaoTang/GEM}.

\end{abstract}

\begin{figure}[t]
  \centering
    \includegraphics[width=\linewidth]{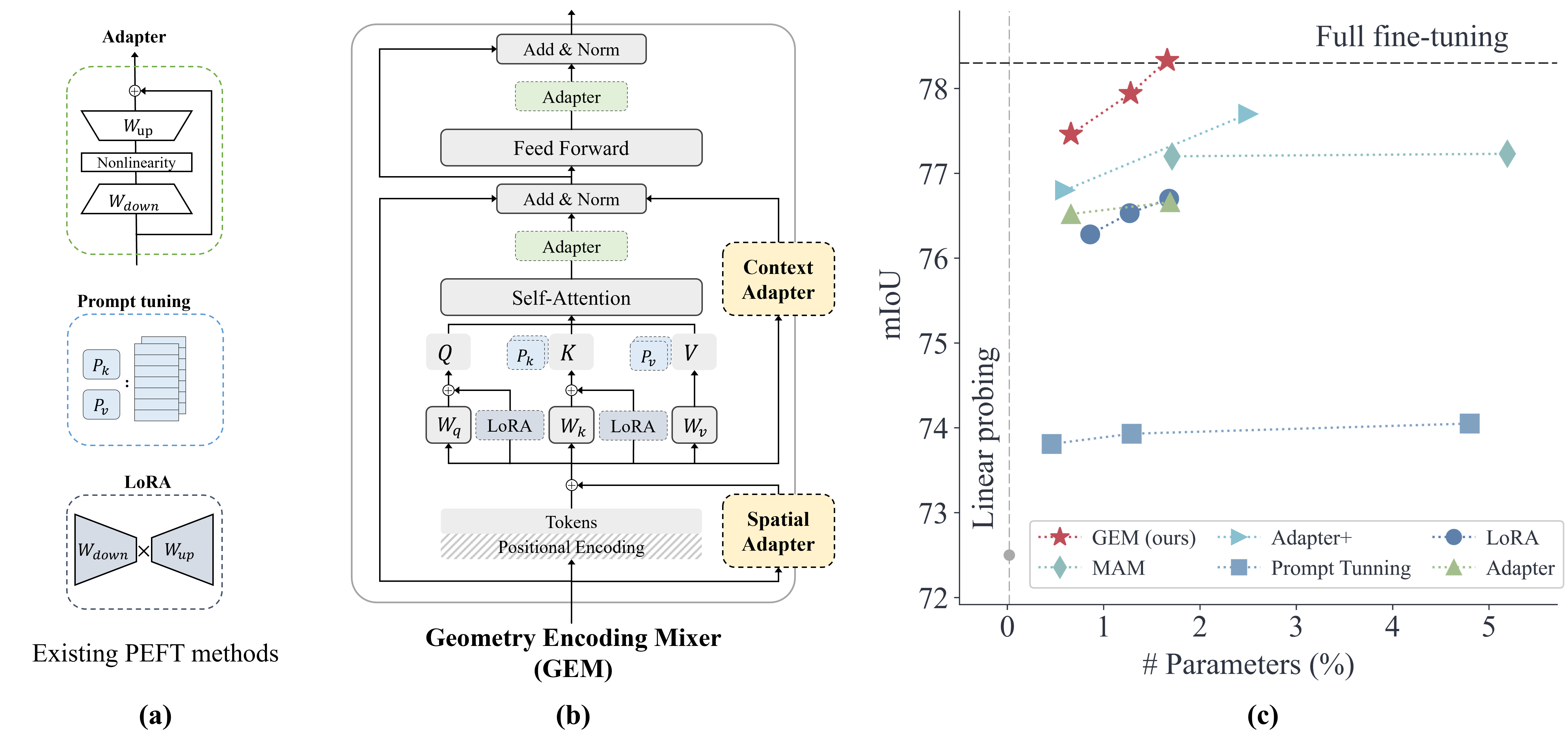}
  % % \hfill
  % \captionsetup{skip=0pt}                                    % Tighten overall caption
  % \captionsetup[subfigure]{aboveskip=0pt, belowskip=-5pt}  % Tighten subcaptions
  % \begin{subfigure}{0.59\textwidth}
  %   \centering
  %   \includegraphics[width=\linewidth]{fig/main_struct.pdf}
  %   \caption{}
  %   \label{fig:main}
  % \end{subfigure}%
  % \hfill%
  % % performance curve
  % \begin{subfigure}{0.39\textwidth}
  %   \centering
  %   \includegraphics[width=\linewidth]{fig/main_curve.png}
  %   \caption{}
  %   \label{fig:main}
  % \end{subfigure}%
  % % \hfill
  % 
  \caption{
    \textbf{(a)} Existing PEFT methods, such as adapters, prompt tuning, and LoRA, focus on adaptations inside attention and feed-forward layers.
    \textbf{(b)} In contrast, Geometry Encoding Mixer (GEM) explicitly encodes the geometric cues and mixes them into the pre-trained model, by the Spatial Adapter refining the pre-trained positional encoding and the Context Adapter complementing the local attention.
    \textbf{(c)} By capturing 3D spatial details and scene-wide geometry context, GEM reaches full fine-tuning performance while tuning $<2\%$ parameters, outperforming existing PEFT methods.
    % Direct application of existing PEFT methods on 3D point cloud yields marginal gains, due to their inability of capturing 3D spatial details and scene-wide geometry context.
    }
  \label{fig:main}
\end{figure}

\section{Introduction}
\label{sec:intro}
Point cloud semantic segmentation is a fundamental task for scene understanding that underpins many real-world applications, from autonomous driving and unmanned aerial vehicles to augmented reality~\cite{ptsurvey, ptsurvey_seg, ptsurvey_tr}.
Recent breakthroughs in large-scale models for language and 2D vision~\cite{vit,dinov2,clip,deepseekv3,qwen25,gpt4o} have spurred an analogous trend toward training powerful point cloud backbones capable of capturing rich semantics from 3D scenes~\cite{sonata,ptv3,ppt}.

Despite significant capabilities, adapting large pre-trained backbone models to specific downstream target domains is non-trivial. The typical strategy heavily relies on full fine-tuning, a process that is both computationally expensive and storage-intensive, since each downstream task demands an independent copy of all model parameters.
%In addition, such large models typically require correspondingly large training datasets to reach their full potential, yet they tend to overfit when fine-tuned on a small target dataset, also known as catastrophic forgetting~\cite{adap_simple,adap_fusion}.
Furthermore, fine-tuning large models typically requires extensive datasets to prevent overfitting and catastrophic forgetting~\cite{adap_simple,adap_fusion}, while large datasets are usually inaccessible when adapted to smaller, specialized tasks.
% While linear probing~\cite{sonata,lin_decaf,tr_mae}, training only a linear classifier given a frozen backbone, could be an effective solution, it generally yields inferior performance compared to full fine-tuning.
To address these issues, parameter-efficient fine-tuning (PEFT) methods offer a promising performance in reducing the cost of adaptation while reaching the performance of full fine-tuning with significantly fewer parameters.

% Representative approaches include
% adapters~\cite{adap}, which insert lightweight bottleneck modules into the backbone;
% LoRA~\cite{lora}, which performs efficient weight updates on attention layer via low-rank matrix decompositions;
% and prompt tuning~\cite{prefix,prefix_prompt}, which prepends trainable embeddings to the input sequence of attentions to learn a form of prompt tokens.

While PEFT has been extensively validated in language processing and 2D vision tasks, its effectiveness on large-scale 3D point clouds remains largely unexplored and inadequately addressed.
% This has thus resulted in an urgent need for parameter-efficient fine-tuning (PEFT) techniques for large-scale 3D point cloud, a challenge that remains largely open
As demonstrated empirically in \cref{fig:main}, existing PEFT methods only yield limited performance gains if directly applied to a 3D point cloud backbone.
% Therefore, we first explore by porting the existing PEFT methods to 3D point cloud backbone.
% We attribute this shortfall to the unique challenges in 3D point cloud data, which yet are overlooked by these methods.
% Unlike language or image, point cloud are essentially unordered set of coordinates in $\mathbb R^3$ with no fixed structure.
% Owing to the intrinsic irregularity and sparsity, point clouds appear to have scene-dependent variations, such as point density and structural patterns, shaped by various sensing protocols and underlying geometries.
%These variations induce pronounced geometric and spatial distribution shifts between large-scale pre-training datasets and the target domains.
% 
We tend to attribute this limitation to a key challenge that is often overlooked by current PEFT methods when using for 3D point cloud: unlike structured data such as text or images, point clouds are inherently unordered sets of coordinates in $\mathbb{R}^3$. They exhibit strong irregularity, sparsity, and structural variability, shaped by different sensing protocols and scene geometries.
These factors lead to significant geometric and spatial distribution shifts between large-scale pre-training datasets and downstream domains, which existing PEFT methods fail to account for.
%However, among the typical PEFT methods, adapters and LoRA adapt the model on a per-point basis and prompt tuning introduce external tokens at the scene level, all disregarding the important spatial and geometric cues during the adaptation for 3D point cloud.
%Moreover, though 3D transformer backbones follow the general design of transformer~\cite{transformer}, they typically employ local attention to avoid the prohibitive computational cost of global attention over a large number of points.
%The large amount of input tokens and the constrained local attention further imped the effectiveness of existing PEFT methods.
In particular, representative PEFT methods, such as LoRA~\cite{lora}, adapters \cite{adap} and prompt tuning \cite{prefix,prefix_prompt}, either adapt models at an isolated, per-point level or insert fixed external tokens at the global level, thus failing to adequately capture important geometric and spatial contexts inherent in 3D scenes.
Moreover, to avoid the prohibitive computational cost of global attention over a large number of points, current 3D transformers predominantly employ local attention mechanisms without explicitly modeling global contexts, which further restricts the potential of current PEFT approaches in performance.

To address the above challenges, we re-examine PEFT for 3D scene understanding and hypothesize that effective PEFT on 3D scenes could take advantage of explicitly modeling both fine-grained local spatial patterns and global geometric contexts.
We propose that neglecting either aspect would degrade fine-tuning performance: local-only adaptations may be prone to noise due to the lack of broader context consideration, whereas global-only methods would miss local details and lose precision. This motivates a novel PEFT adaptation framework tailored specifically for 3D scenes to bridge these local and global scales.

We introduce Geometry Encoding Mixer (GEM), a geometry-aware module for parameter-efficient fine-tuning of point cloud transformer on 3D scenes.
It comprises two complementary components: a spatial adapter at local neighborhood and a context adapter capturing the scene context.
In the spatial adapter, we employ a lightweight 3D convolutional bottleneck that operates on points neighborhood, enriching the pre-trained positional encoding by learning fine-grained local spatial details at target domains.
In the context adapter, we introduce a set of learned latent tokens to serve as global context vectors.
These latent tokens interact with the full point cloud through efficient attention, forming a bottleneck at the token dimension and bypassing the constraints of local attention, and thus aggregating scene-specific context from across the entire point cloud.
By fusing these two paths, the Geometry Encoding Mixer effectively bridges local and global representations, providing the fine-tuned model with a richer understanding of the 3D scene than either component alone.

Empirically, we validate our approach on large-scale 3D scene datasets, including both indoor~\cite{scannet,scannet200,s3dis,scannetpp} and outdoor~\cite{semkitti} scenes.
Our results indicate that models equipped with GEM consistently achieve performance matching or sometimes surpassing full fine-tuning methods, updating merely \textasciitilde 1.6\% of model parameters, while being fewer than existing PEFT alternatives.
These findings underscore the importance of explicitly modeling geometric and spatial contexts for efficient and effective adaptation in 3D scene understanding.
To the best of our knowledge, our work represents the first exploration and validation of PEFT approaches tailored explicitly for large point cloud transformer under large-scale 3D scenes, hoping to establish a foundation for future research and practical deployments.

\section{Related Work}
\label{sec:related_work}

\paragraph{Point cloud segmentation.}
Point clouds serve as an efficient representation for large-scale 3D scenes.
Consequently, point cloud segmentation becomes a fundamental task that drives the design of 3D backbone architectures.
Since the seminal work PointNet~\cite{pointnet,pointnet++}, numerous 3D backbone architectures have been proposed for this task. These backbones either project the points onto a grid-like structure, \eg 3D voxel grid, to exploit 3D convolutional networks~\cite{minkowski, ocnn, seg_vx_SEGCloud, occuseg, submanifold, seg_vx_ScanComplete}, or directly process the raw unordered point sets~\cite{dgcnn, pointcnn, pointconv, randlanet, kpconv, closerlook}.
Recently, inspired by the success of large transformer models in natural language and 2D vision~\cite{transformer, vit,tr_detr, tr_mask2former}, researchers have started training increasingly large and powerful point cloud backbones~\cite{pointnext, pttransformer, octformer, cdformer, stratified, ptv2, ptv3}.
% Although these models achieve strong performance, they typically need to be trained from scratch for each new dataset and often require specialized training recipes to reach optimal results.
Although these models achieve strong performance, they typically need to be trained from scratch for each new dataset and often remain data-hungry~\cite{weak_10few,weak_erda,weak_erda_ext,weak_otoc}. 
To reach optimal results, specialized training recipes and learning targets are also commonly required and actively explored~\cite{bound_cbl,seg_vx_ScanComplete,pointnext}.

Following the paradigm of large-scale pre-training that has proven successful in vision and language domains~\cite{clip, sam, dinov2, deepseekv3, gpt4o, tr_mae}, researchers have begun exploring similar strategies for 3D point clouds.
In particular, self-supervised learning (SSL) on large 3D scene datasets has demonstrated promising results~\cite{pointcontrast, p4contrast, contrastscene, msc}.
However, most such efforts focus on modest convolutional backbones like SparseUNet~\cite{minkowski} rather than transformer-based architectures.
Only recently has an SSL approach been applied to a transformer-based 3D backbone: Sonata~\cite{sonata} introduced SSL for a large point transformer encoder~\cite{ptv3}, but even this method still requires full fine-tuning on each downstream dataset to achieve optimal performance.

In this work, we explore effective PEFT methods for large pre-trained point cloud backbones, with the goal of reducing the computational and storage overhead when adapting them to downstream datasets, while matching the performance of full fine-tuning.

\paragraph{Parameter-efficient fine-tuning (PEFT).}
% To alleviate the burden of training when adapting to downstream tasks and datasets, there have been early attempts of adding extra parameters and selectively tuning partial models for CNNs~\cite{adap_cnn}. For recent large transformer backbones and foundation models~\cite{vit,sam,llama}, the full fine-tuning have become ever more infeasible due to their high demands on memory and computational resources.
% In this regards, PEFT methods have emerged to allow efficient fine-tuning and deployment of large foundation models.
The ever-growing size of transformer-based foundation models\cite{vit,sam,llama} makes full fine-tuning for downstream task prohibitively expensive in both memory and computation.
PEFT methods address this challenge by adapting large models while updating only a small fraction of their parameters.
Existing approaches fall into four broad categories.
% PEFT methods can be broadly categorized into the following four strategies.

% Selective tuning methods focus on adapting the model by tuning only a small set of the existing model parameters, such as the classifier head in linear probing~\cite{dinov2,lin_decaf}, biases~\cite{select_bitfit}, or parameters with large gradients~\cite{select_gps}.
Selective fine-tuning methods update a carefully chosen subset of the original weights. One simple variant is linear probing~\cite{dinov2,lin_decaf} that trains only the classification head. Other strategies restrict training to specific parts of the network, \eg, tuning only bias terms~\cite{select_bitfit} and selecting a subset of parameters based on gradient magnitude criteria~\cite{select_gps}.

% Adapter-based tuning, originally proposed for convolutional networks~\cite{adap_cnn}, introduces small trainable modules into a frozen backbone.
% This idea has been extensively developed for transformers in NLP, with various adapter architectures proposed. These adapter modules are typically implemented as bottleneck MLP layers with residual connections, inserted either sequentially~\cite{adap,adap+} or in parallel within transformer blocks~\cite{adap_mam,adap_ciat,adap_bias,adap_former}.
Adapter-based tuning inserts lightweight bottleneck modules into an otherwise frozen backbone.
First explored for CNNs~\cite{adap_cnn} and later extended to transformers, adapters may be placed sequentially~\cite{adap,adap+} or in parallel~\cite{adap_mam,adap_ciat,adap_bias,adap_former} to the original modules.
Beyond single task, adapters can be composed or fused, promoting knowledge sharing without catastrophic forgetting~\cite{adap_adamix,adap_fusion}.

Prompt-based tuning techniques, including prompt tuning~\cite{prefix_prompt} and prefix tuning~\cite{prefix}, extend to the prompting paradigm in large language models~\cite{prefix_promptsurvey}.
Instead of hand-crafted prompts, these methods learn task-specific vectors that are prepended to the input sequence~\cite{prefix_visualprompt,prefix_prompt} or injected as extra tokens in transformer layers~\cite{prefix,prefix_promptv2,prefix_vpt}, thereby steering the model without modifying its original weights \cite{prefix_prompt,prefix_promptsurvey}.

% Low-rank adaptation (LoRA) constrains the weight updates to a low-dimensional subspace. The pioneering work~\cite{lora} approximate the effect of full fine-tuning by learning the low-rank decomposition matrices for the weight updates in attention layers.
Low-rank adaptation (LoRA) constrains the weight updates to a low-dimensional subspace by learning a pair of low-rank matrices that are added to each pre-trained weight, typically in the attention layers~\cite{lora}.
This design approximates the effect of full fine-tuning while introducing only a small number of additional parameters.
Successors further enhance the approximation ability, robustness, or rank allocation~\cite{lora_rs,lora_dora,lora_ada,lora_reft,ia3}. 

Recent works also propose hybrid schemes that highlight common design principles~\cite{adap_llmsurvey}, for instance, integrating their respective strengths under a unified adapter framework~\cite{adap_mam}.

Although PEFT has proved effective in language and 2D vision, a direct transfer to 3D scene understanding is often sub-optimal, largely due to the unordered nature of point clouds and the prominence of geometric cues.
Our work thus investigates PEFT strategies tailored to large-scale 3D scenes.

\paragraph{PEFT for 3D point cloud.}
In the context of 3D point clouds, PEFT remains relatively underexplored.
Existing methods thus far have focused mostly on object-level inputs with limited spatial scale.
% IDPT~\cite{ptpeft_instance} propose to leverage graph network~\cite{dgcnn} to dynamically generate prompts for each input point cloud.
% DAPT~\cite{ptpeft_dyadap} further modulates the scale of the dynamically generated prompts and apply learnable affine transformation to regulate the features.
% Zhang \et~\cite{ptpeft_posprompt} propose to generate prompt embedding from the center point of input patches and further combine with adapters to emphasizes the importance of positional encoding.
Some approaches introduce prompt tokens that adapt to 3D data, where the prefix tokens are dynamically generated from intermediate features~\cite{ptpeft_instance, ptpeft_dyadap, ptpeft_gaprompt} or from the spatial centers of local patches~\cite{ptpeft_posprompt}.
% PointGST~\cite{ptpeft_gst} leverages graph network in spectual domain to provide the backbone with required global geometry pattern.
% STAG~\cite{ptpeft_stag} introduces a side graph network for efficient fine-tuning on point cloud object classification.
Several works also introduce auxiliary side networks, operating either in the spectral domain~\cite{ptpeft_gst} or the spatial domain~\cite{ptpeft_stag, ptpeft_pointlora}, to provide geometric context.
% Point-PEFT~\cite{ptpeft_peft} utilizes domain dataset to create a bank of prompts and then finetune with adapters.
% Sun \et~\cite{ptpeft_prompt} combines prompt-tuning on language encoder with adapter after point cloud encoder to explore efficiently tuning for open-vocabulary point cloud understanding.
Other methods construct banks of prompt vectors using domain-specific dataset to inject 3D prior knowledge~\cite{ptpeft_peft}, or combine language-side prompt tuning with point-cloud adapters to handle open-vocabulary recognition~\cite{ptpeft_prompt}.

% Compared with mostly plain transformer backbones on object-level point cloud with a few thousand input points~\cite{shapenetpart,partnet}, transformer models for 3D scene requires additional concerns for large-scale input points that generally takes more than million points~\cite{scannet,s3dis} and can be significantly resource-intensive to train.
% We thus suggest that PEFT methods on 3D scene urges more exploration and would benefit the fine-tuning of large 3D scene models greatly.
Compared to object-level datasets~\cite{shapenetpart,partnet}, 3D scene understanding involves much larger inputs containing on the order of millions of points~\cite{scannet,s3dis}.
This scale amplifies computational challenges and underscores the need for PEFT techniques expressly designed for large 3D scene models, which is however a research direction that remains largely unexplored.
% Developing PEFT methods tailored to large 3D scene models is therefore a promising and largely untapped research direction.

\section{Methodology}
\label{sec:method}

\begin{figure}[t]
  \centering
  \captionsetup{skip=0pt}                                    % Tighten overall caption
  \includegraphics[width=.8\linewidth]{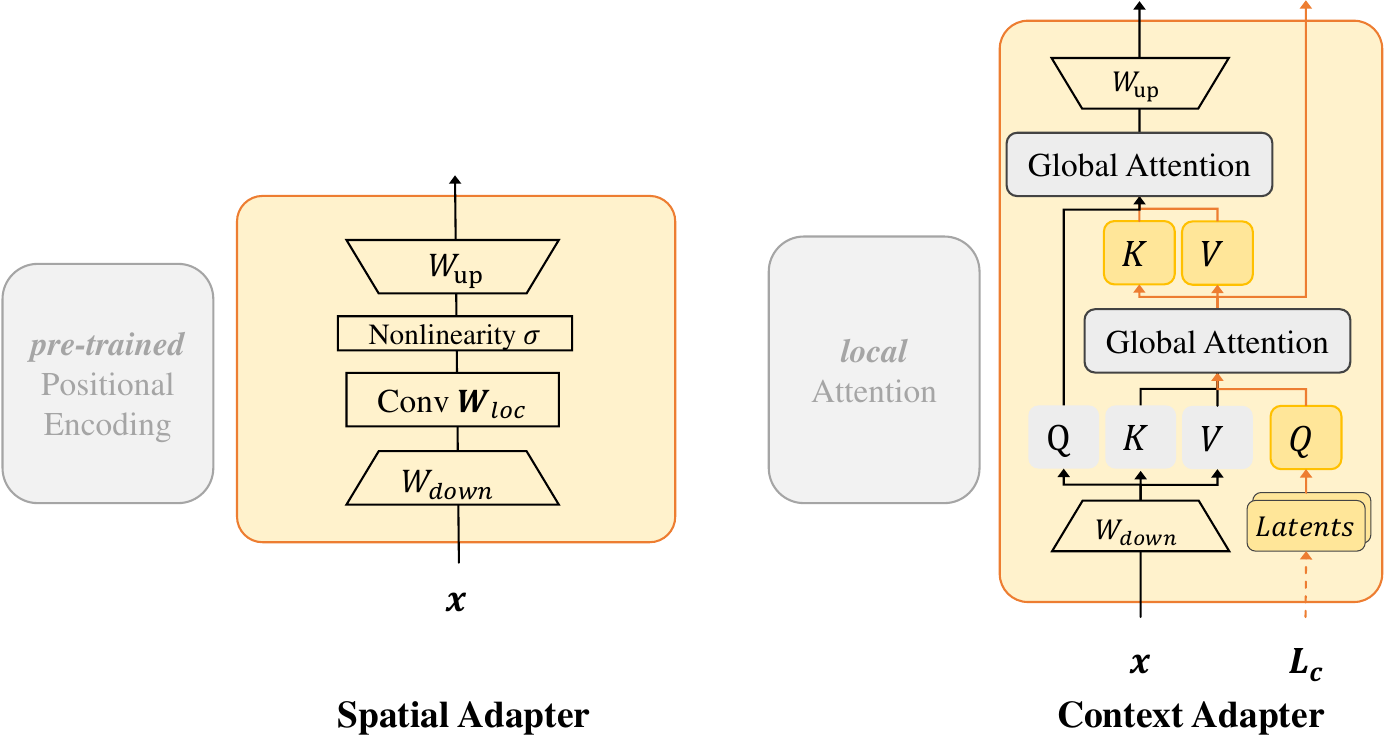}
  \caption{
    \textbf{Geometry Encoding Mixed.} We propose the spatial adapter to enhance the pre-trained positional encoding, and the context adapter to overcome the local attention mechanism, thus enhancing the efficient adaptation on large-scale 3D scenes with explicit geometry encoding.
  }
  \label{fig:struct}
\end{figure}

We propose the \emph{Geometry Encoding Mixer} (GEM) as a lightweight parameter-efficient module for fine-tuning point-cloud transformer. \cref{fig:struct} summarises the overall architecture.

In particular, GEM consists of two complementary adaptations: a \emph{Spatial Adapter} that refines the positional encoding of each point, injecting local geometry information overlooked by generic adapters; and a \emph{Context Adapter} that distills a compact set of latent tokens that broadcast global scene-level cues.
Both components follow the residual, bottleneck design of classic adapters, yet emphasize and operate on spatial rather than channel space.

% Formally, let $\mathbf{X}\in\mathbb{R}^{N\times d}$ be the input features (for $N$ points, $d$-dimensional). GEM produces adapted features
% Drawing on the unified view of adapters~\cite{adap_mam,adap_llmsurvey}, GEM can be viewed as a 3D-specific form of prompt tuning that sidesteps the prohibitive cost of global attention on large point sets.

\subsection{Preliminaries}
We first consider the direct application of existing PEFT methods on current point-cloud transformers.

\paragraph{Challenges under 3D.}
Following transformer~\cite{transformer, vit}, point-cloud transformers also build upon self-attention layer (Attn) followed by feed-forward network (FFN).
However, due to the large scale input points at \emph{million} scale, global self-attention over all points would demand prohibitively large GPU memory and compute resource, if ever possible, due to the quadratic complexity of the attention operations: $\mathcal{O}(\text{Attn})=n^2$, where $n$ denotes the number of input points.
In this regard, state-of-the-art point-cloud transformers propose various designs of local attentions~\cite{pttransformer,ptv2,ptv3,stratified,octformer} to cap the complexity, where each point can only attend to points within the same patch. With a patch size of $p$, the complexity of local attention is then $\mathcal{O}(\text{Attn}_\text{loc})=np$. For example, it is set to $p=16$ in PT~\cite{pttransformer} and $p=1024$ in the larger PTv3~\cite{ptv3}.
% To further reduce the complexity, point cloud backbones generally take a hierarchical process to gradually down-sample the point cloud.
% For example, PTv3~\cite{ptv3} serialize the unordered points into ordered sequence based on space filling curve and perform local attention in each patch of 1024 points.

The irregular and sparse nature of 3D point cloud, together with these architecture designs, greatly impede the application of standard PEFT methods such as adapter, LoRA, and prompt tuning.
As shown in \cref{fig:main}, directly applying existing methods yields only marginal gains.
We attribute this to their lack of spatial structure awareness of 3D geometry and the inability to capture scene-wide context under the local attention regimes of modern point cloud backbones.

\paragraph{Adapters.} The adapter-based methods~\cite{adap,adap+} insert small modules after the attention or FFN layers:
\begin{equation}
  f_\text{adapter}(\bm x) = \bm x + \sigma(\bm x \bm W_\text{down}) \bm W_\text{up},
\end{equation}
where $\bm W_\text{down}\in \mathbb R^{d\times r}$ and $\bm W_\text{up}\in \mathbb R^{r\times d}$ are down and up projections that form a bottleneck structure with $r \ll d$ and activation $\sigma$ in the mid, and are also surrounded by residual connection.
As generic MLP bottlenecks, adapters adapt the model on a per-point basis, overlooking the important spatial cues in 3D point cloud.
% In addition, the frozen local attentions would limit the adapted point features in capturing the global scene.

\paragraph{LoRA.} LoRA methods~\cite{lora} typically adapt the attention projection layers and approximate the full fine-tuning with a bottleneck in weight space, transforming the attention into:
\begin{equation}
  \text{Attn}_\text{LoRA} = \text{Attn}(\bm Q + \Delta \bm Q, \bm K+ \Delta \bm K, \bm V),
\end{equation}
where $\Delta \bm Q = \bm x \bm W_\text{down} \bm W_\text{up}$, with $\bm W_\text{down}\in \mathbb R^{d\times r}$ and $\bm W_\text{up}\in \mathbb R^{r\times d}$, and similarly for $\Delta \bm K$ with a different pair of weights.
While LoRA can leverage the strong ability of the attention layers, we note that all attentions in point cloud transformer are constrained within local patches.
LoRA methods are thus inherently limited by the prevalent design of local attention. With only limited rank $r << d$, it could fail to provide global context of the target dataset.

\paragraph{Pompt tuning.} Prompt tuning methods~\cite{prefix,prefix_prompt} explicitly introduce global tokens into the attention:
\begin{equation}
  \text{Attn}_\text{prompt} = \text{Attn}(\bm Q, [\bm P_K; \bm K], [\bm P_V; \bm V]),
\end{equation}
where $\bm P_K, \bm P_V\in\mathbb R^{m\times d}$ are two sets of tokens prepending to the original key-value pair.
In spite of the constraint of local patches, prompt tokens can be duplicated into each patch.
However, with $m << n$, it can be hard for a few static external tokens to capture scene-specific context that can vary across point clouds within the same dataset.
In addition, it also overlooks the spatial pattern of the point cloud, lacking the adaptation to local point features.
%  as it also focus on the adaptation of attention modules and is thus restricted by the design of local attention.

\paragraph{Others.} There are more methods that explore other aspects. Selective tunings propose to tune a small subset of the frozen model parameters, such as BitFit~\cite{select_bitfit} that tunes only bias terms.
Hybrid methods are also explored, such as MAM~\cite{adap_mam} that couples FFN adapter with prompt tuning for attention layer.
These methods still inherit the constraints from those representative PEFT methods, which are hindered by the local attention and fail to recognize the spatial structure.

In summary, existing methods overlook 3D geometry and remain confined to local patches, motivating a PEFT design that incorporates spatial awareness with scene-wide context, introduced next.

\subsection{Our Methodology: Geometry Encoding Mixer (GEM)}
% We develop our method to address these limitations.
To address these limitations, we present the Geometry Encoding Mixer (GEM).
It refines positional encodings and injects scene context for efficient adaptation on large-scale 3D scenes:
\begin{align}
  \bm{x} &\leftarrow \bm{x} + \text{pos}(\bm{x}) + f_{\text{spatial}}(\bm{x}),\\
  \bm{x} &\leftarrow \bm{x} + \text{Attn}_{\text{loc}}(\bm{x}) + f_{\text{context}}(\bm{x}),
\end{align}
where $\bm{x}$ denotes input points and $\text{pos}(\cdot)$ is the pre-trained positional embedding. \cref{fig:struct} illustrats the overall structure.

\paragraph{Spatial Adapter (SA).}
Point clouds are sparse, irregular samples in 3D. Adapting to their fine-grained geometry requires explicitly modeling local neighborhoods, which prior PEFT methods fail to address.

To this end, the spatial adapter refines per-point positional encoding through a lightweight 3D convolutional bottleneck.
Concretely, for each point $\bm x$, we consider a 3D grid and gather points in the vicinity voxels as neighbors, denoted by $\mathcal N$:
\begin{equation}
  f_\text{spatial}(\bm x) = \bm x + \sigma( \sum_{i\in\mathcal N} \bm W^i_\text{loc} (\bm x \bm W_\text{down})) \bm W_\text{up},
\end{equation}
where $\bm W_\text{down}\in \mathbb R^{d\times r}$ and $\bm W_\text{up}\in \mathbb R^{r\times d}$ are down and up projectsion with $r \ll d$. $\bm W^i_\text{loc}\in \mathbb R^{r\times r}$ composes the spatial kernel weights for $i$-th neighboring voxel and $\sigma(\cdot)$ is the nonlinearity, such as ReLU.

With a common kernel dimension $k=3$, the spatial adapter touches at most $k^3$ neighbours per point and adds $2rd + k^3r^2$ parameters, rendering a complexity of $\mathcal O(2ndr + nk^3r^2) = \mathcal O(nd)$ given $k, r \ll n, d$.
It thus functions as an efficient convolution-based positional encoding~\cite{pos_cpe,pos_conv} in parallel to the pre-trained positional encoding, thereby capturing fine spatial layout to match the target distribution.

\paragraph{Context Adapter (CA).}
% In face of the large-scale input point cloud at million scale, point cloud transformers generally apply local self-attention within patches of size $p$ to avoid the quadratic cost of global attention~\cite{pttransformer,ptv3,sonata}.
Due to the local attention, any local and channel-wise adaptation would be constrained from acquiring scene-wide context, hindering the adaptation performance on downstream tasks like semantic segmentation.

To inject global context without breaking the $\mathcal O(n^2)$ barrier, we introduce $m$ latent tokens $\bm L \in \mathbb{R}^{m\times r}$, where $m \ll n$, that attend to all points once:
\begin{align}
\bm{L}_c &= \text{Attn}\bigl(\bm{L}\bm{W}^Q,\,
                                  \bm{K},\,
                                  \bm{V}\bigr),\\
f_\text{context}(\bm x)  &= \bm{x} \;+\;
               \text{Attn}\bigl(\bm{Q},\,
                                \bm{L}_c\bm{W}^K,\,
                                \bm{L}_c\bm{W}^V\bigr) \bm W_\text{up},
\end{align}
where  $\bm L \in\mathbb R^{m\times r}$ and $\bm Q, \bm K, \bm V = \bm x \bm W_\text{down}^Q, \bm x \bm W_\text{down}^K, \bm x \bm W_\text{down}^V$ are down-projected point features.

As both attention costs $O(nm)$ with $m \ll n$, global aggregation is affordable and at the same level of complexity as the conventional prompt tuning~\cite{prefix,prefix_prompt}.
Furthermore, we update $\bm{L} \leftarrow \bm{L}+\bm{L}_c$ at each adapter insertion to share across layers, yielding dynamic prompts that capture the context of current scene, unlike static prefixes in prompt tuning.

\paragraph{Discussion.}
% As also shown in~\cref{fig:main}, they still obtain sub-optimal performance, appealing the urgent need of PEFT methods tailored for large-scale 3D scenes.
% 
% The standard PEFT methods create channel bottleneck to remain efficient by reducing the large feature dimensions~\cite{lora,adap}, we explicitly create spatial bottleneck.
% In this view, the spatial adapter offers a spatial bottleneck of $3\times3\times3$ in the form of convolution at local neighborhoods, and scene context adapter forms a bottleneck of $m$ over the whole point cloud.
% Therefore, we are able to efficiently model the local and global structure of point cloud without operating on the full amount of $n$ points.
% 
% Although individual ingredients, convolutional positional encodings~\cite{pos_cpe,ptv3,octformer,pos_cvt} and attentions in latent feature space~\cite{tr_perceiver,tr_set,tr_groundingdino} have been similarly proposed, they mainly consider full training with additional design for 2D vision tasks.
% We are, to the best of our knowledge, the first to motivate their effective application as PEFT methods for 3D scenes.
% Our results demonstrate that explicitly adapting geometric cues is essential for effective fine-tuning of point-cloud transformers.
% 
In comparison to existing PEFT methods, GEM explicitly biases the adaptation to 3D geometry and context.
The spatial adapter captures localized variations that generic adapters, such as LoRA and basic adapters, would miss, since those ignore spatial structure and use the positional encodings from pre-training.
The context adapter circumvents the local attention design to provide scene-wide context for efficient adaptation with few parameters.
% Empirically, this dual design closes the gap between full fine-tuning and previous PEFT baselines on challenging point cloud benchmarks while tuning fewer than $2\%$ of parameters.
By combining these two components, our approach enables scene-aware adaptation of 3D transformers, where each point is tuned with respect to both its neighbors and the entire point cloud.
It thus results in significantly improved performance on 3D scene datasets with only a lightweight adaptation overhead\footnote{
We study the empirical overheads in the supplementary \cref{sec:supp_impl}.
}.

%%%%%%%%%%%%%%%%%%%%%%%%%%%%%%%%%%%%%%%%%%%%%%%%%%%%%%%%%%%%%%%%%%%%%%%%%%%%%%%%%%%%%%%%%%%%%%%%%%%%%%%%%%%%%%%%%%%%%%%%

%%%%%%%%%%%%%%%%%%%%%%%%%%%%%%%%%%%%%%%%%%%%%%%%%%%%%%%%%%%%%%%%%%%%%%%%%%%%%%%%%%%%%%%%%%%%%%%%%%%%%%%%%%%%%%%%%%%%%%%%

\section{Experiments}
\label{sec:exp}

We present the results of our proposed \method on semantic segmentation of large-scale 3D scene datasets and experiment with both self-supervised pre-trained backbone, Sonata~\cite{sonata}, and supervised pretrained one, PTv3-PPT~\cite{ppt}, for investigation\footnote{
For more generalization and comparisons in other settings, such as convolutional model and 3D shape analysis, please refer to the supplementary \cref{sec:supp_exp}.
}.
We also provide ablation studies to reveal the detailed effects of different components better.

\subsection{Experimental Setup}
We primarily follow the comprehensive protocols proposed in Sonata~\cite{sonata}, covering ScanNet~\cite{scannet}, ScanNet200~\cite{scannet200}, ScanNet++~\cite{scannetpp} and S3DIS~\cite{s3dis}.

We adopt two representative pre-trained backbones, the Sonata model~\cite{sonata} with large-scale self-supervised training and the PTv3-PPT~\cite{ppt} with supervised pre-training on multiple large-scale curated datasets.
We compare \method with the existing popular PEFT methods, including bias-tuning from BitFit~\cite{select_bitfit}, Adapter~\cite{adap}, LoRA~\cite{lora}, and Prompt Tuning~\cite{prefix}. We consider linear probing as a baseline and the full fine-tuned model as our target strong reference.
In addition, we note that the Sonata (full.) specifically introduces a task-specific decoder for semantic segmentation during the fine-tuning, which may not reflect the fair comparison. Therefore, we revive to the standard full fine-tuning without additional decoder network, which is denoted by Sonata (ft.).

% Full fine-tuning and training from scratch generally provide upper and lower bounds.
For training, we follow the widely accepted fine-tuning setups to update only the inserted or selected weights with the pre-trained backbone weights remaining frozen.
All PEFT baselines follow the implementations from released code, adopt the suggested common practice~\cite{code_peft,code_adapters}, and are tuned to their best validation setting in \cref{fig:main}(c).
% For a fair comparison, we follow the common practice in implementing the existing PEFT methods~\cite{code_peft,code_adapters} and adopt the default setting as the one when each method reaches its best performance in \cref{fig:main}.
Specifically, we set the default rank to be $r=32$ and the number of learnable tokens to be $m=4$.
More details are given in the supplementary.

\subsection{Performance Comparison}

\begin{table*}
    \centering
    \resizebox{\linewidth}{!}{%
        % \tablestyle{2.6pt}{1.08}
        \tablestyle{1pt}{1.05}
        \begin{tabular}{=l+r+r+c+c+c+c+c+c+c+c+c+c+c+c+c+c+c}% lrrccccccccccccccc
\toprule
Semantic Seg. &\multicolumn{2}{c}{Params} &\multicolumn{3}{c}{ScanNet Val~\cite{scannet}} &\multicolumn{3}{c}{ScanNet200 Val~\cite{scannet200}} &\multicolumn{3}{c}{ScanNet++ Val~\cite{scannetpp}} &\multicolumn{3}{c}{S3DIS Area 5~\cite{s3dis}} &\multicolumn{3}{c}{S3DIS 6-fold~\cite{s3dis}} \\
\cmidrule(lr){1-1} \cmidrule(lr){2-3} \cmidrule(lr){4-6} \cmidrule(lr){7-9} \cmidrule(lr){10-12} \cmidrule(lr){13-15} \cmidrule(lr){16-18}
Methods &\multicolumn{1}{c}{Learn.} &\multicolumn{1}{c}{Pct.} &mIoU &mAcc &allAcc   &mIoU &mAcc &allAcc     &mIoU &mAcc &allAcc     &mIoU &mAcc &allAcc     &mIoU &mAcc &allAcc \\
\cmidrule(lr){1-18}
\multicolumn{18}{l}{\emph{Training from scratch}} \\
\specialrule{0em}{1pt}{1pt}

% \rowstyle{\color{darkgray}}
SparseUNet~\color{defaultcolor}{\cite{minkowski}}   & 39.2M     & 100\%     & 72.3  & 80.2  & 90.0  & 25.0  & 32.9  & 80.4  & 28.8  & 38.4  & 80.1  & 66.3  & 72.5  & 89.8  & 72.4  & 80.9  & 89.9  \\

% \rowstyle{\color{darkgray}}
PTv3~\color{defaultcolor}{\cite{ptv3}}              & 124.8M    & 100\%     & 77.6  & 85.0  & 92.0  & 35.3  & 46.0  & 83.4  & 42.1  & 53.4  & 85.6  & 73.4  & 78.9  & 91.7  & 77.7  & 85.3  & 91.5  \\

\cmidrule(lr){1-18}
\multicolumn{18}{l}{\emph{Full fine-tuning}} \\
\specialrule{0em}{1pt}{1pt}

% \rowstyle{\color{darkgray}}
Sonata (full.)~\color{defaultcolor}{\cite{sonata}}  & 124.8M    & 100\%     & 79.4 & 86.1 & 92.5    & 36.8 & 46.5 & 84.4    & 43.7 & 55.8 & 86.6    & 76.0 & 81.6 & 93.0    & 82.3 & 89.9 & 93.3 \\

% \rowstyle{\color{darkgray}}
Sonata (ft.)~\color{defaultcolor}{}    & 108.5M     & 100\%                             & 78.3 & 85.9 & 92.3    & 37.3 & 47.8 & 83.7    & 49.8 & 61.2 & 87.6    & 72.4 & 79.0 & 92.2    & 79.5 & 87.3 & 92.3 \\
\cmidrule(lr){1-18}
\multicolumn{18}{l}{\emph{PEFT methods}} \\
\specialrule{0em}{1pt}{1pt}

Sonata (lin.)                           & 0.02M     & 0.02\%    & 72.5 & 83.1 & 89.7 & 29.3 & 41.6 & 81.2 & 37.3 & 50.9 & 84.3 & 72.3 & 81.2 & 90.9 & 76.5 & 87.4 & 90.8 \\

% basic
\ \ \plus BitFit~\cite{select_bitfit}   & 0.2M      & 0.2\%     & 74.7 & 84.7 & 90.8    & 32.5 & 45.5 & 82.0    & 42.4 & 56.5 & 85.7    & 73.9 & 82.0 & \second{91.5}     & 76.1 & 87.1 & 91.0 \\

\ \ \plus Adapter~\cite{adap}           & 2.8M      & 2.5\%     & \second{77.0} & 85.4 & \second{91.8}     & \second{33.6} & \second{45.7} & 82.5   & 42.6 & 57.5 & 85.6      & 73.8 & 82.9 & 91.5    & 76.4 & 87.5 & \second{91.4} \\

\ \ \plus LoRA~\cite{lora}              &  1.9M      & 1.7\%     & 76.7 & \second{85.7} & 91.7    & 33.6 & 45.5 & \second{82.7}     & \second{44.2} & \second{58.0} & \best{86.5}    & \second{74.5} & \best{83.2} & 91.5        & \second{77.4} & \second{87.8} & 91.2 \\

\ \ \plus Prompt Tunning~\cite{prefix}  & 5.5M      & 4.8\%     & 74.3 & 84.1 & 90.5    & 31.4 & 44.4 & 81.6     & 41.2 & 56.2 & 84.8      & 73.4 & 82.5 & 91.0        & 73.7 & 86.5 & 90.5 \\

% improvement - 
% \ \ \plus MAM~\cite{adap_mam}           & 5.9M      & 5.2\%    &  &  & &  & & & & & & & & & & & \\
% 77.23     5.19%  -

% \ \ \plus Adapter+          &           &           &  &  & &  & & & & & & & & & & & \\

\ \ \plus \best{\method (ours)}                & 1.8M      & 1.6\%     & \best{78.3} & \best{86.6} & \best{92.3}    & \best{35.6} & \best{46.9} & \best{83.3}         & \best{46.6} & \best{60.3} & \second{86.3}     & \best{75.1} & \second{83.0} & \best{92.1}    & \best{77.9} & \best{88.2} & \best{92.1} \\

% Sonata (dec.)                           & 16.3M     & 13\%      & 79.1 & 86.6 & 92.7 & 33.5 & 44.5 & 84.1 & 40.9 & 52.6 & 86.3 & 74.5 & 80.4 & 92.6 & 81.5 & 88.8 & 93.0 \\

\bottomrule
\end{tabular}

% TODO:
% decoder cfg

    }% resize
    \vspace{-3mm}
    \caption{
      \textbf{Semantic segmentation.}
      % Method with $\dagger$ introduce task-specific head during the fine-tuning.
    }
    \label{tbl:main}
    \vspace{1mm}
    % \vspace{-6mm}
\end{table*}

\paragraph{Main results.}
\cref{tbl:main} shows that \method consistently surpasses all representative PEFT methods across datasets.
It matches the performance of full fine-tuning on most datasets and even exceeds it on ScanNet++\cite{scannetpp}.
ScanNet++ comprises large and diverse scenes captured at sub-millimeter resolution, diverging greatly from the spatial distribution of common pre-training datasets such as ScanNet~\cite{scannet}.
Under this domain gap, the joint modeling of local spatial cues and global scene context introduced by \method proves especially beneficial.

Interestingly, LoRA~\cite{lora} and adapter~\cite{adap} deliver similar scores, despite acting on different transformer components.
This observation implies that, when updates remain local and geometric is not modeled explicitly, the precise choice of adaptation target can be secondary.
In addition, prompt tuning~\cite{prefix} underperforms even the linear probing baseline in the S3DIS 6-fold evaluation, revealing the penalty of ignoring spatial structure during fine-tuning.

\paragraph{Data efficiency.}
We assess PEFT methods under the scenarios of limited data and annotations in \cref{tbl:scannet_eff}.
The results demonstrate the exceptional data efficiency of \method, which outperforms both other PEFT methods and the full fine-tuning counterparts.
Notably, under extreme data scarcity, such as 1\% of the labeled scenes and limited annotations (20 points per scene), \method surpasses both Sonata (full.) and Sonata (ft.), highlighting its superiority in low-data regimes.

\paragraph{Supervised pre-training.}
While self-supervised methods dominate the language domain, supervised pre-training remains competitive for vision~\cite{sam}.
To probe the limits of PEFT under large-scale 3D supervision, we adopt recent advances in 3D supervised pre-training~\cite{ppt}, which achieve leading performance by training on a large collection of curated, labeled scene datasets.
As shown in \cref{tbl:supervised}, existing PEFT methods can even degrade performance, likely suffering from negative transfer~\cite{nt_deep,nt_avoid}.
In contrast, \method improves upon supervised backbone, further outperforming the larger PTv3~\cite{ptv3} with only 1.6\% additional parameters.

\paragraph{PEFT with decoder.}
A dedicated segmentation decoder (13\% of the total parameters) is known to lift fully fine-tuned baselines.  
We therefore ask whether PEFT can still add value in the presence of such a task-specific head.
\cref{tbl:decoder} answers in the affirmative: \method outperforms all competing PEFT variants and even surpasses its own fully fine-tuned counterpart, setting a new state-of-the-art on ScanNet.  
The result underscores the merit of jointly modeling local geometry and scene-level context, even when the decoder already provides task-specific capacity.

\paragraph{Outdoor segmentations.}
During the original evaluation of Sonata~\cite{sonata}, it employs separate pre-trained models for indoor and outdoor scenarios, due to the significant domain gaps between indoor and outdoor scenes.
% Specifically, the model conducts pre-training on indoor datasts for indoor benchmarks and pre-training on outdoor datasets for outdoor benchmarks.
% It thus leaves an open question: \emph{can model pre-trained on indoor datasets generalize to outdoor scenarios?}
Here, we test a stronger setting: transferring an indoor pre-trained backbone to an outdoor autonomous-driving benchmark.
We replace the input layer with a randomly initialized counterpart to match dimensionality and freeze all remaining weights. 

\cref{tbl:semseg_outdoor} shows that \method narrows the gap to a model trained from scratch on outdoor data, yet a noticeable performance gap persists.
Closing this gap remains an interesting direction for future work.
% As shown in \cref{tbl:semseg_outdoor}, the proposed \method largely closes the gap towards the performance of model that is trained from scratch.
% However, there is still a noticeable gap between the performance obtained by pre-training and fine-tuning model on outdoor datasets, appealing for future exploration.

%##################################################################################################
\begin{table}[t]
\RawFloats
% \hfill
\resizebox{\linewidth}{!}{%
\begin{minipage}[t]{.55\textwidth}
\vspace{0pt}
\centering
%#################################################
% data eff
%#################################################
\resizebox{\linewidth}{!}{%
  \tablestyle{1pt}{1.05}
  \begin{tabular}{lrrrrrrrrrrr}\toprule
Data Efficiency &\multicolumn{5}{c}{Limited Scenes (Pct.)} &\multicolumn{5}{c}{Limited Annotation (Pts.)} \\
\cmidrule(lr){1-1} \cmidrule(lr){2-6} \cmidrule(lr){7-11}
Methods                                 & 1\%~ & 5\%~ & 10\%~ & 20\%~ & Full~                 &~20~ &50~ &100~ &200~ &Full~ \\
\midrule
% \cmidrule(lr){1-11}
\multicolumn{11}{l}{\emph{Training from scratch}} \\
\specialrule{0em}{1pt}{1pt}

SparseUNet~\cite{minkowski} &26.0 &47.8 &56.7 &62.9 &\darkgray{72.2} &41.9 &53.9 &62.2 &65.5 &\darkgray{72.2} \\

% PTv2~\cite{ptv2} &24.8 &48.1 &59.8 &66.3 &\darkgray{75.4} &58.4 &66.1 &70.3 &71.2 &\darkgray{75.4} \\

PTv3~\cite{ptv3} &25.8 &48.9 &61.0 &67.0 &\darkgray{77.2} &60.1 &67.9 &71.4 &72.7 &\darkgray{77.2} \\

\cmidrule(lr){1-11}
\multicolumn{11}{l}{\emph{Full fine-tuning}} \\
\specialrule{0em}{1pt}{1pt}

Sonata (full)~\cite{sonata}             & 45.3 & 65.7 & 72.4 & 72.8 &\darkgray{79.4} &~70.5 & 73.6 & 76.0 & 77.0 &\darkgray{79.4} \\
Sonata (ft.)                            & 44.4 & 63.2 & 71.3 & 72.3 &\darkgray{78.3}    & 69.6 & 72.6 & 75.3 & 76.2 &\darkgray{78.3}   \\

\cmidrule(lr){1-11}
\multicolumn{11}{l}{\emph{PEFT methods}} \\
\specialrule{0em}{1pt}{1pt}
%                                       1sc     5sc     10sc    20sc                    20pt   50pt   100pt  200pt    
Sonata (lin.)                           & 43.6 & 62.5 & 68.6 & 69.8 &\darkgray{72.5}    & 69.0 & 70.5 & 71.1 & 71.5 &\darkgray{72.5} \\

% basic
\ \ \plus BitFit~\cite{select_bitfit}   & 46.5 & \second{64.9} & 69.9 & 71.7 &\darkgray{74.7}    & 71.0 & 72.3 & 72.9 & 73.6 &\darkgray{74.7} &   \\

\ \ \plus Adapter~\cite{adap}           & 46.4 & 64.2 & \second{70.1} & 72.2 &\darkgray{77.0}    & 71.8 & 73.4 & 74.7 & 75.0 &\darkgray{77.0} &   \\

\ \ \plus LoRA~\cite{lora}              & \second{46.6} & 63.0 & 70.1 & \second{72.6} &\darkgray{76.7}    & \second{72.1} & \second{73.6} & \second{75.2} & \second{75.5} &\darkgray{76.7} &   \\

\ \ \plus Prompt Tunning~\cite{prefix}  & 45.5 & 62.6 & 68.9 & 71.1 &\darkgray{74.3}    & 70.2 & 71.5 & 72.5 & 72.8 &\darkgray{74.3} &   \\

\ \ \plus \best{\method (ours)}                & \best{47.5} & \best{65.6} & \best{71.0} & \best{73.3} &\darkgray{78.3}    & \best{72.3} & \best{74.7} & \best{76.2} & \best{76.6} &\darkgray{78.3} &   \\

% Sonata (dec.)                         & 44.5 & 64.1 & 69.8 & 72.5 &\darkgray{79.1} & 69.8 & 73.1 & 75.0 & 76.3 &\darkgray{79.1} \\
\bottomrule
\end{tabular}

}%
\caption{
  \textbf{Data efficiency.}
}
\label{tbl:scannet_eff}
\end{minipage}
\hfill
% \hspace{.1em}
\begin{minipage}[t]{.445\textwidth}
\vspace{0pt}
\centering
%#################################################
% PPT
%#################################################
\resizebox{\linewidth}{!}{%
  \tablestyle{1pt}{1.05}
  \begin{tabular}{lrrccc}%
\toprule
Supervised Pre-train. &\multicolumn{2}{c}{Params} &\multicolumn{3}{c}{ScanNet Val~\cite{scannet}} \\
\cmidrule(lr){1-1} \cmidrule(lr){2-3} \cmidrule(lr){4-6}
Methods &\multicolumn{1}{c}{Learn.} &\multicolumn{1}{c}{Pct.} &mIoU &mAcc &allAcc    \\
\cmidrule(lr){1-6}
\multicolumn{6}{l}{\emph{Training from scratch}} \\
\specialrule{0em}{1pt}{1pt}

% \rowstyle{\color{darkgray}}
SparseUNet~\color{defaultcolor}{\cite{minkowski}}   & 39.2M     & 100\%                 & 72.3  & 80.2  & 90.0      \\

% \rowstyle{\color{darkgray}}
PTv3~\color{defaultcolor}{\cite{ptv3}}              & 124.8M    & 100\%                 & 77.6  & 85.0  & 92.0      \\

\cmidrule(lr){1-6}
\multicolumn{6}{l}{\emph{Full fine-tuning}} \\
\specialrule{0em}{1pt}{1pt}

% \rowstyle{\color{darkgray}}
% Sonata (full.)~\cite{sonata}                        & 124.8M    & 100\%     & 79.4 & 86.1 & 92.5        \\

PTv3-PPT (ft.)~\cite{ppt}                           & 97.4M     & 100\%     & 78.6 & 86.0 & 92.5        \\

\cmidrule(lr){1-6}
\multicolumn{6}{l}{\emph{PEFT methods}} \\
\specialrule{0em}{1pt}{1pt}

PTv3-PPT (lin.)                                     & 0.1M      & 0.1\%     & \second{78.6} & 85.9 & \second{92.5}             \\

% basic
\ \ \plus BitFit~\cite{select_bitfit}               & 0.2M      & 0.2\%     & 78.2 & 85.6 & 92.3                    \\

\ \ \plus Adapter~\cite{adap}                       & 1.8M      & 1.8\%     & 78.5 & 85.9 & 92.4           \\

\ \ \plus LoRA~\cite{lora}                          & 1.4M      & 1.7\%    & 78.4 & \second{86.0} & 92.4                     \\

\ \ \plus Prompt Tunning~\cite{prefix}              & 4.8M      & 4.8\%     & 78.3 & 85.9 & 92.4                    \\

\ \ \best{\plus \method (ours)}                     & 1.8M      & 1.6\%     & \best{79.1} & \best{86.6} & \best{92.6}      \\

% Sonata (dec.)                           & 16.3M     & 13\%      & 79.1 & 86.6 & 92.7 & 33.5 & 44.5 & 84.1 & 40.9 & 52.6 & 86.3 & 74.5 & 80.4 & 92.6 & 81.5 & 88.8 & 93.0 \\

\bottomrule
\end{tabular}
}%
\vspace{0pt}
% \vspace{14pt}
\caption{
  \textbf{Supervised pre-training.}
}
\label{tbl:supervised}
\end{minipage}
% \hfill
}% resizebox
\end{table}
%##################################################################################################

%##################################################################################################
\begin{table}[t]
\RawFloats
% \hfill
\resizebox{\linewidth}{!}{%
\begin{minipage}[t]{.493\textwidth}
\vspace{0pt}
\centering
%#################################################
% tight budget
%#################################################
\resizebox{\linewidth}{!}{%
  \tablestyle{1pt}{1.05}
  \begin{tabular}{lrrccc}
\toprule
with Decoder &\multicolumn{2}{c}{Params} &\multicolumn{3}{c}{ScanNet Val~\cite{scannet}}  \\
\cmidrule(lr){1-1} \cmidrule(lr){2-3} \cmidrule(lr){4-6}
Methods &\multicolumn{1}{c}{Learn.} &\multicolumn{1}{c}{Pct.} &mIoU &mAcc &allAcc \\

\cmidrule(lr){1-6}
\multicolumn{6}{l}{\emph{Training from scratch}} \\
\specialrule{0em}{1pt}{1pt}

PTv3~\color{defaultcolor}{\cite{ptv3}}  & 124.8M    & 100\%     & 77.6 & 85.0 & 92.0    \\

\cmidrule(lr){1-6}
\multicolumn{6}{l}{\emph{Full fine-tuning}} \\
\specialrule{0em}{1pt}{1pt}

% \rowstyle{\color{darkgray}}
Sonata (full.)~\cite{sonata}            & 124.8M    & 100\%     & 79.4 & 86.1 & 92.5    \\
Sonata (full.)$^*$                      & 124.8M    & 100\%     & 78.5 & 86.3 & 92.4    \\

% % \rowstyle{\color{darkgray}}
% Sonata (ft.)                            & 108.5M    & 100\%     & 78.3 & 85.9 & 92.3    \\

\cmidrule(lr){1-6}
\multicolumn{4}{l}{\emph{PEFT methods with decoder}} \\
\specialrule{0em}{1pt}{1pt}

Sonata (dec.)                           & ~16.3M    & 13.1\%    & \second 79.1 & 86.6 & \second 92.7    \\

Sonata (dec.)$^*$                       & 16.3M     & 13.1\%    & 77.2 & 85.9 & 91.9    \\

% basic
\ \ \plus BitFit~\cite{select_bitfit}   & 16.4M      & 13.2\%     & 78.1 & 86.1 & 92.3   \\

\ \ \plus Adapter~\cite{adap}           & 19.1M     & 15.0\%    & 78.2 & 86.5 & 92.3    \\

\ \ \plus LoRA~\cite{lora}              & 18.2M     & 14.3\%    & 78.9 & \second 87.0 & 92.5     \\

\ \ \plus Prompt Tunning~\cite{prefix}  & 22.6M     & 17.2\%     & 77.4 & 86.1 & 92.1    \\

% improvement - 
% \ \ \plus MAM~\cite{adap_mam}           & 5.9M      & 5.2\%    &  &  & &  & & & & & & & & & & & \\

% \ \ \plus Adapter+          &           &           &  &  & &  & & & & & & & & & & & \\

\ \ \plus \best{\method (ours)}         & 18.1M     & 14.3\%     & \best{79.5} & \best{87.3} & \best{92.7}    \\
\bottomrule
\end{tabular}
}%
\vspace{0pt}
% \vspace{14pt}
\caption{
  \textbf{PEFT with segmentation decoder.} Methods with $^*$ reports our re-produced results.
}
\label{tbl:decoder}
\end{minipage}
% \hfill
\hspace{.1em}
\begin{minipage}[t]{.49\textwidth}
\vspace{0pt}
\centering
%#################################################
% decoder
%#################################################
\resizebox{\linewidth}{!}{%
  \tablestyle{1pt}{1.05}
  \begin{tabular}{lrrccc}
\toprule
Outdoor Seg. &\multicolumn{2}{c}{Params} &\multicolumn{3}{c}{Sem.KITTI Val~\cite{semkitti}}  \\
\cmidrule(lr){1-1} \cmidrule(lr){2-3} \cmidrule(lr){4-6}
Methods &\multicolumn{1}{c}{Learn.} &\multicolumn{1}{c}{Pct.} &mIoU &mAcc &allAcc \\

\cmidrule(lr){1-6}
\multicolumn{6}{l}{\emph{Training from scratch}} \\
\specialrule{0em}{1pt}{1pt}

PTv3~\color{defaultcolor}{\cite{ptv3}}  & 124.8M    & 100\% & 69.1 & 76.1 & 92.6    \\

\cmidrule(lr){1-6}
\multicolumn{6}{l}{\emph{Full fine-tuning}} \\
\specialrule{0em}{1pt}{1pt}

% \rowstyle{\color{darkgray}}
Sonata (full.)~\cite{sonata}$^\dagger$  & 124.8M    & 100\% &   72.6 & 77.9 & 93.4      \\

% \rowstyle{\color{darkgray}}
Sonata (ft.)                            & 108.5M    & 100\% &   68.8 & 75.2 & 92.6      \\

\cmidrule(lr){1-6}
\multicolumn{6}{l}{\emph{PEFT methods}} \\
\specialrule{0em}{1pt}{1pt}

Sonata (lin.)$^\dagger$                 & 0.02M     & 0.02\%    & 62.0 & 72.5 & 91.0    \\
Sonata (lin.)                           & 0.02M     & 0.02\%    & 52.0 & 63.3 & 87.1    \\

% basic
\ \ \plus BitFit~\cite{select_bitfit}   & 0.2M      & 0.2\%     & 59.9 & 70.6 & 91.0    \\

\ \ \plus Adapter~\cite{adap}           & 2.8M      & 2.5\%     & 62.5 & 72.2 & 92.0    \\

\ \ \plus LoRA~\cite{lora}              & 1.9M      & 1.7\%     & \second 63.8 & \second 74.0 & \second 92.0    \\

\ \ \plus Prompt Tunning~\cite{prefix}  & 5.5M      & 4.8\%     & 59.1 & 71.0 & 90.8    \\

% improvement - 
% \ \ \plus MAM~\cite{adap_mam}           & 5.9M      & 5.2\%    &  &  & &  & & & & & & & & & & & \\

% \ \ \plus Adapter+          &           &           &  &  & &  & & & & & & & & & & & \\

\ \ \plus \best{\method (ours)}         & 1.8M      & 1.6\%     & \best{67.7} & \best{75.5} & \best{93.1}    \\

\bottomrule
\end{tabular}
}%
\caption{
  \textbf{Outdoor semantic segmentation.} Methods with $^\dagger$ use outdoor datasets for pre-training.
}
\label{tbl:semseg_outdoor}
\end{minipage}
% \hfill
}% resizebox
\end{table}
%##################################################################################################

\subsection{Ablations and Analsys}
We mainly consider the linear probing, Sonata (lin.), as baseline and ablate on ScanNet~\cite{scannet} to better investigate the key factors for effective PEFT in 3D scenes, shown in \cref{tbl:ablations}.
For more studies on the effect of available parameters, please refer to the supplement.
% For more studies on hyper-parameters, please refer to the supplement

\paragraph{Individual effectiveness of SA and CA.}
To validate our initial hypothesis, we study the individual effectiveness of the SA and CA in \cref{tbl:abl_comp}.
We find that both adapters can already be superior to the baseline and competitive to the existing PEFT methods.
We notice that the SA takes the majority part of the introduced parameters, which is partially due to the inherently expensive 3D convolution with $k^3$ kernels.
Such challenge is also 3D specific and urges us to develop the CA, rather than solely relying on the positional encoding to capture 3D geometry.
By combining the two form of spatial adapters, we obtain \method that reaches the best performance by comprehensively exploring the 3D scene while remaining parameter-efficient.

\paragraph{Understanding the latent tokens.}
In \cref{tbl:abl_tokens}, we are motivated to further investigate how efficient and effective the proposed CA could be. We thus study how few tokens it could afford before failing to capture the global context.
Surprisingly, we find that CA can use as few as one single token to effectively express the global scene context, obtaining clear improvement over SA-only approach.
We consider this to be the effectiveness of weight decomposition, where we are akin to approximating the $n^2$ global self-attention weights with two $n\times 1$ matrices. In comparison to LoRA that decomposes in the weight space, we decompose the spatial attention weight matrices and form a spatial bottleneck at token dimension.
In addition, while we are aware that related topics have been similarly explored for efficient attentions~\cite{tr_perceiver,tr_set}, we are, to the best of our knowledge, the first to motivate their effective application as PEFT methods for 3D scenes understanding.

% We are, to the best of our knowledge, the first to motivate their effective application as PEFT methods for 3D scenes.
% Our results demonstrate that explicitly adapting geometric cues is essential for effective fine-tuning of point-cloud transformers.

\paragraph{Sharing the latent tokens.}
The last experiment, \cref{tbl:abl_dist} probes whether the latent tokens learned by CA should remain layer-local or be shared across the hierarchy, assessing its ability to capture dynamic and scene-specific context.
When each layer keeps its own bank of tokens (\textit{N/A}), \method already surpasses SA-only, confirming that CA supplies complementary global cues.
Re-using the same tokens within each encoder stage (\textit{per-stage}) yields a further yet modest gain, suggesting that a coherent context inside each resolution level helps the network reconcile local geometry with mid-range structure.

The strongest result emerges when a single set of latent tokens is shared by \emph{all} transformer blocks.
We attribute this improvement to its consistency with the fact that 3D point clouds represent the same underlying 3D geometry in spite of their irregularity and sparsity.
From this perspective, such globally shared latent tokens confine the model to adapt to the actual underlying scene, rather than overfitting on irrelevant patterns due to the noisy sampled point cloud.

\paragraph{Revealing the geometry cues in latent tokens.}
Under the above assumption that latent tokens regularize the model to capture the underlying 3D scene, we visualize the attention weights and compare them to those produced by other PEFT methods that also inject global context, such as prompt tuning.

\begin{figure}[t]
  \centering
  % \hfill
  \captionsetup{skip=-10pt}                                    % Tighten overall caption
  \includegraphics[width=\linewidth]{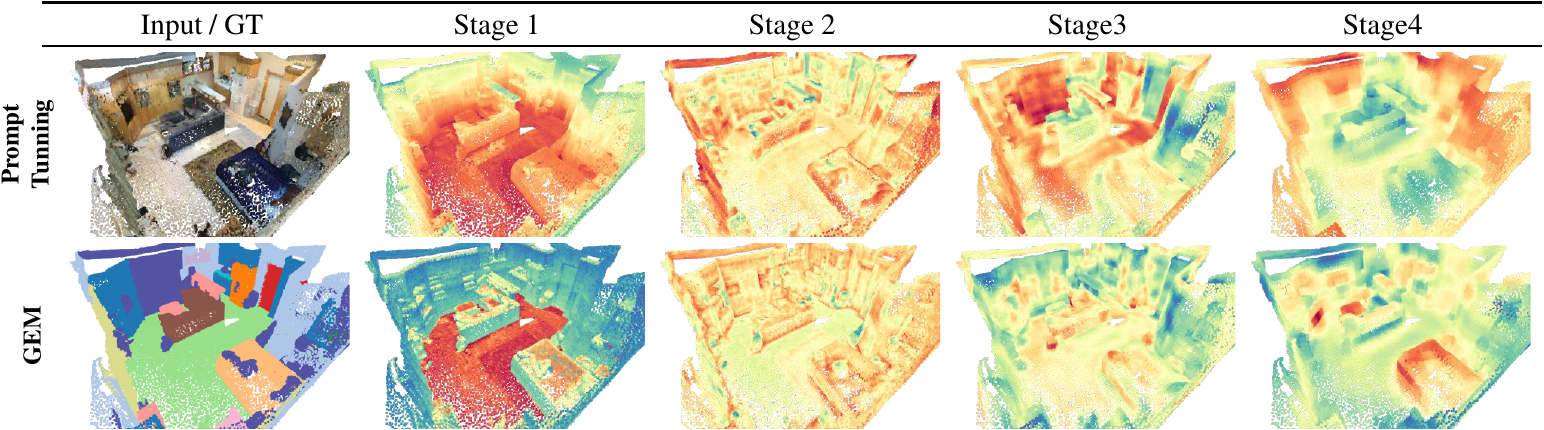}
  \vspace{-.5em}
  \caption{
    We visualize the attention weights of our latent tokens, comparing to the attentional weights with the prompt tuning, showing the enhanced geometry cues produced by \method. More visualizations in \cref{sec:supp_vis}.
  }
  \label{fig:attn}
\end{figure}

As shown in \cref{fig:attn}, starting from the first stage, our latent tokens learn to produce crisper attention weights that follow geometric cues, such as the simple geometric primitives of floor surface.
In deeper stages, the tokens progressively attend to more salient geometric objects, including clutter on the table and sofa.
In contrast, the static tokens introduced by prompt tuning produce yield blurrier attention patterns that often span across object and region boundaries.

These observations suggest that the proposed latent tokens effectively capture and broadcast scene-wide geometric context.
Combined with the local spatial structure extracted by 3D convolutions, this leads to superior performance, establishing our method as an effective PEFT approach for 3D scene understanding.

%##################################################################################################
% ablations
\begin{table*}[t]
\vspace{-.2em}
\centering
\resizebox{\linewidth}{!}{%
%#################################################
% SA/CA
%#################################################
\subfloat[
\textbf{SA} and \textbf{CA} compose of effective PEFT for 3D scenes, individually and jointly.
\label{tbl:abl_comp}
]{
\begin{minipage}{0.38\linewidth}{
\vspace{0pt}
\begin{center}
\tablestyle{1pt}{1.05}
% \resizebox{1.05\linewidth}{!}{%
\begin{tabular}{l c c c c}
% $\lambda$  & $0$ & $1$ \\
\toprule
& \textbf{SA}~ & ~\textbf{CA}~ & Params & ~mIoU \\
\cmidrule{1-5}
Sonata (lin.)  &   &   &0.02\%  & 72.5  \\
\multirow{3}{*}{\textbf{+ \method}}
  & \checkmark  &   & 1.0\%   & 77.2 \\
% sonata_sft_adapter-ppos-hdim32-lin-subm3/Log_2025-04-21_11-53-47
  & & \checkmark     & 0.6\%   & 77.3 \\
% sonata_sft_alattn8-pattsc-hdim32-mh2/Log_2025-04-30_08-45-14
% sonata_sft_alattn4-pattsc-hdim32-mh2/Log_2025-04-27_19-26-44

  & \baseline \checkmark  & \baseline \checkmark  & \baseline 1.6\% & \baseline 78.3 \\

\end{tabular}
% }% resizebox
\end{center}
}\end{minipage}
}
%#################################################
% latents
%#################################################
% \hfill
\hspace{.8em}
\subfloat[
\textbf{\method} is robust to available latent tokens.
\label{tbl:abl_tokens}
]{
\begin{minipage}{0.3\linewidth}{
\vspace{0pt}
\begin{center}
\tablestyle{1pt}{1.05}
\begin{tabular}{lccc}
\toprule
  $m$  ~ & mIoU & mAcc & allAcc \\
\cmidrule{1-4}
  1   & 78.1 & 86.4 & 92.1 \\
% results/scannet/sonata_sft_adapter-pos-hdim32-lin-subm3_alattn1-pattsc-lstage-hdim32-mh2-lprj/Log_2025-05-16_01-25-51/log_val.txt_best
  \baseline{4}   & \baseline{~78.3~} & \baseline{~86.6~} & \baseline{~92.3~} \\
  8   & 78.2 & 86.8 & 92.3 \\
% results/scannet/sonata_sft_adapter-pos-hdim32-lin-subm3_alattn4-pattsc-lglb-ldim64-hdim32-mh2-lprj/Log_2025-05-10_07-33-00/log_val.txt_best <<<
  
  \\
\end{tabular}  
\end{center}
}\end{minipage}
}
%#################################################
% sharing latents
%#################################################
% \hfill
\hspace{.8em}
\subfloat[
\textbf{\method} can provide better capabilities when sharing latent tokens.
\label{tbl:abl_dist}
]{
\begin{minipage}{0.32\linewidth}{
\vspace{0pt}
\begin{center}
\tablestyle{1pt}{1.05}
% \resizebox{1.05\linewidth}{!}{%
\begin{tabular}{l c c c }  
% $\lambda$  & $0$ & $1$ \\
% \diagbox[height=1.5em]{$L_\text{DA}$}{$\lambda$}  & $0$ & $1$ \\
\toprule
~ strategy ~  & mIoU & mAcc &  allAcc  \\
\cmidrule{1-4}
N/A.          & 77.8 & 86.1 &  92.1  \\
% results/scannet/sonata_sft_adapter-pos-hdim16-lin-subm3_alattn4-pattsc-hdim32-mh2-lprj     mIoU=77.81 mAcc=86.07 OA
per-stage.    & 78.0 & 86.1 &  92.0  \\
\baseline global  & \baseline 78.3 & \baseline 86.6 & \baseline 92.3  \\
\\
\end{tabular}

% }% resizebox
\end{center}
}\end{minipage}
}
% \hfill
}
%#################################################
\vspace{-.5em}
\caption{Ablations on \method. If not specified, the backbone is the self-supervised pre-trained Sonata~\cite{sonata} with no additional task-specific decoder, evaluated on ScanNet~\cite{scannet}.
Default settings are marked in \colorbox{baselinecolor}{gray}.}
\label{tbl:ablations}
% \vspace{-1.5em}
\end{table*}
%##################################################################################################
%##################################################################################################

\section{Conclusion}
This study presents GEM, a geometry-aware fine-tuning module tailored for large-scale 3D scene segmentation. By combining local spatial refinement and global context modeling, GEM addresses key challenges inherent in adapting pre-trained 3D models to new domains. Experimental results across multiple benchmarks confirm its ability to achieve competitive performance with minimal parameter updates. These findings highlight the value of incorporating geometric priors into efficient model adaptation and offer a promising direction for scalable deployment in 3D scene understanding tasks.

\paragraph{Limitation and future work.}
While \method achieves promising performance with little overhead in complexity, we notice it relies on the assumption that the model can be fully parallelized. For example, it would incur noticeable overhead if using large batch sizes, as the device is forced to process the input sequentially, in spite of the parallel design in \method.
In addition, we are able to successfully train our PEFT methods with a fraction of epochs than the full fine-tuning, \eg 100 rather than 800 epochs, we realize that the gradients need to be back-propagated into early layers and could hinder the training efficiency, especially with dense point clouds like S3DIS~\cite{s3dis}. We thus suggest that it is required a systematic examination on the additional overhead that PEFT methods would experience in 3D scene understanding.

\paragraph{Acknowledgement.}
Dr. Tao's research is partially supported by NTU RSR and Start Up Grants.

\appendix
%%%%%%%%%%%%%%%%%%%%%%%%%%%%%%%%%%%%%%%%%%%%%%%%%%%%%%%%%%%%%%%%%%%%%%%%%%%%%%%%%%%%%%%%%%%%%%%%%%%%%%%%%%%%%%%%%%%%%%%%
% \clearpage
\FloatBarrier
% \clearpage
% \newpage

\section*{Supplementary}
In this supplementary, we provide more materials as follows, \\
% In this supplementary material, we provide more details regarding implementation details in
% Appendix B, more analysis of ERDA in Appendix C, full experimental results in Appendix D, studies
% on parameters in Appendix E, and more visualization in Appendix F.
\cref{sec:supp_impl}~ details the implementation, training, and inference; \\
\cref{sec:supp_exp}~ presents additional experiments and analyses with tight parameter budgets \\
  % \item[\cref{sec:supp_abl}]    for further ablations and analysis on \method;
\cref{sec:supp_vis}~ offers more qualitative visualizations.

\vspace{-0.3em}
\section{Details of Implementation, Training, and Inference}
\label{sec:supp_impl}
Our implementation is based on the open-source codebase Pointcept (\href{https://github.com/Pointcept/Pointcept}{here}) and follows the official implementations for Sonata~\cite{sonata}, PPT~\cite{ppt}, as well as PTv3~\cite{ptv3}.

Leveraging the parameter efficiency of our method, we train on a single 4090 GPU for much fewer epochs to obtain the reported performance. For example, we train on ScanNet for 100 epochs, in contrast to the 800 epochs in the released Sonata configuration.
Our code is available at \href{https://github.com/LiyaoTang/GEM}{https://github.com/LiyaoTang/GEM}.

\cref{tbl:profiling} summarizes empirical latency and memory usage. We intentionally refrain from deployment-specific optimizations such as LoRA weight merging. Although the global attention and local convolution modules introduce extra cost, the runtime and memory footprint stay within the same order of magnitude as other PEFT baselines.

\vspace{-0.3em}
\section{More Experiments and Analysis}
\label{sec:supp_exp}

In spite of the promising results shown in \cref{sec:method}, we suggest that it can better demonstrate the full potential of PEFT methods under broader and more challenging settings.

\paragraph{PEFT for shape analysis.}
While \method is motivated by PEFT for large-scale 3D scenes, most existing 3D PEFT methods target object-level understanding, as discussed in \cref{sec:related_work}.
To assess cross-domain generalization and compare against these existing 3D-specific PEFT methods, we further evaluate \method on 3D shape datasets, ShapeNetPart~\cite{shapenetpart}. As shown in \cref{tbl:partseg_shapenetpart}, \method remains competitive and achieves the best performance. We also observe that common backbones for shape analysis, such as ReCon~\cite{shape_recon}, attach large MLP heads that account for approximately 20\% of all parameters, rendering fine-tuning inefficient.
To broaden the applicability of backbone models for 3D shapes, an interesting direction for future work is to develop architectures with more parameter-efficient heads that are better suited to fine-tuning.

\paragraph{PEFT for 3D convolutional networks.}
While our primary experiments apply \method to state-of-the-art backbones, mainly transformer models, we also verify its generality on convolutional architectures.
In particular, we integrate \method into SparseUNet~\cite{minkowski}, a 3D convolutional model, following the MSC~\cite{msc} protocol to pre-train on ScanNet~\cite{scannet} and fine-tune on ScanNet200~\cite{scannet200}.
As reported in \cref{tbl:semseg_conv}, \method improves performance by +4.6 mIoU over linear probing and outperforms other PEFT baselines, demonstrating robustness beyond transformer models.
We notice the gains are narrower than those on transformer backbones, largely due to the limited capacity of convolutions, as also indicated by a near-collapse linear probing performance in this setting.

\paragraph{PEFT under tight budgets.}
To stress parameter efficiency, we constrain all methods to the same limited learnable-parameter budgets, including enforcing rank $r=1$, allowing 0.1\% parameters, and 1\% parameters.
As reported in \cref{tbl:semseg_tight}, \method delivers consistently higher accuracy, whereas several baselines fail when the budget becomes extremely stringent.

\section{More Visualizations}
\label{sec:supp_vis}
\begin{figure}[t]
  \centering
  \includegraphics[width=\linewidth]{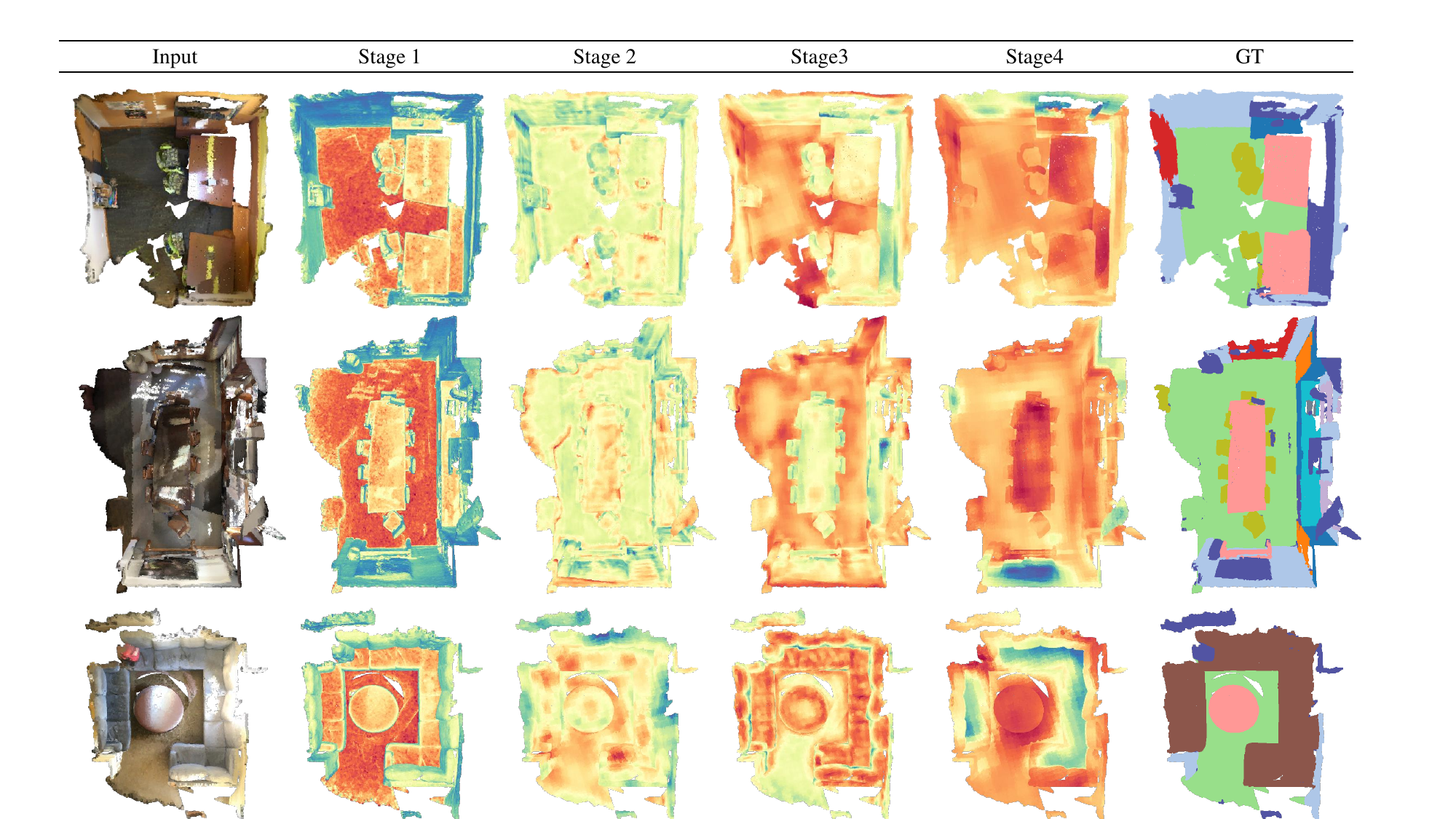}
  \caption{
    Attention maps of our latent tokens in context adapter.
    }
  \label{fig:attn-more}
\end{figure}

\begin{figure}[t]
\centering

\begin{subfigure}{\linewidth}
    \includegraphics[width=\linewidth]{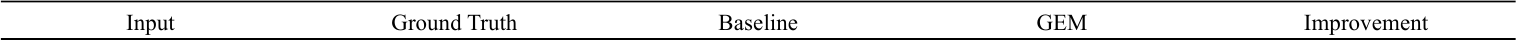}
    \includegraphics[width=\linewidth]{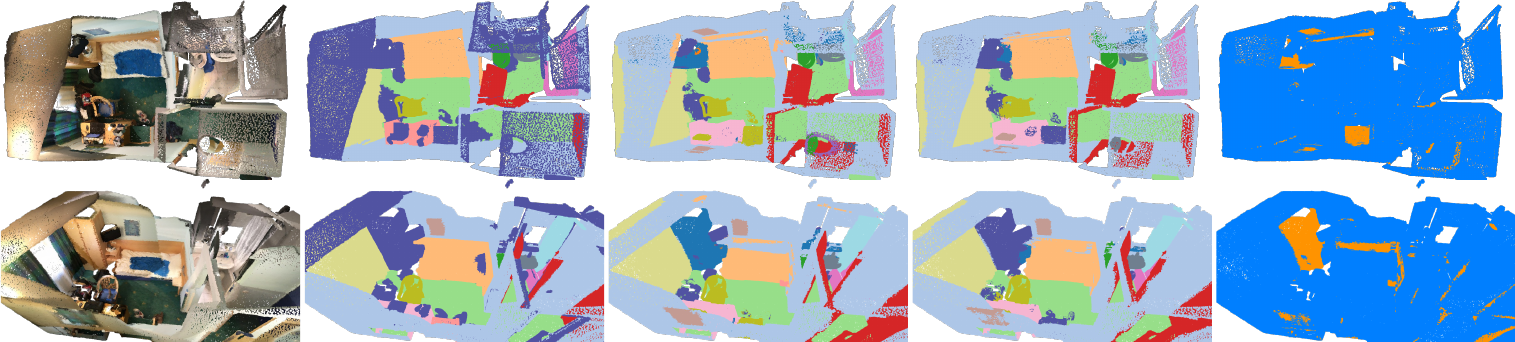}
\caption{Apartments.}
\label{fig:demo-apart}
\end{subfigure}

\begin{subfigure}{\linewidth}
    \includegraphics[width=\linewidth]{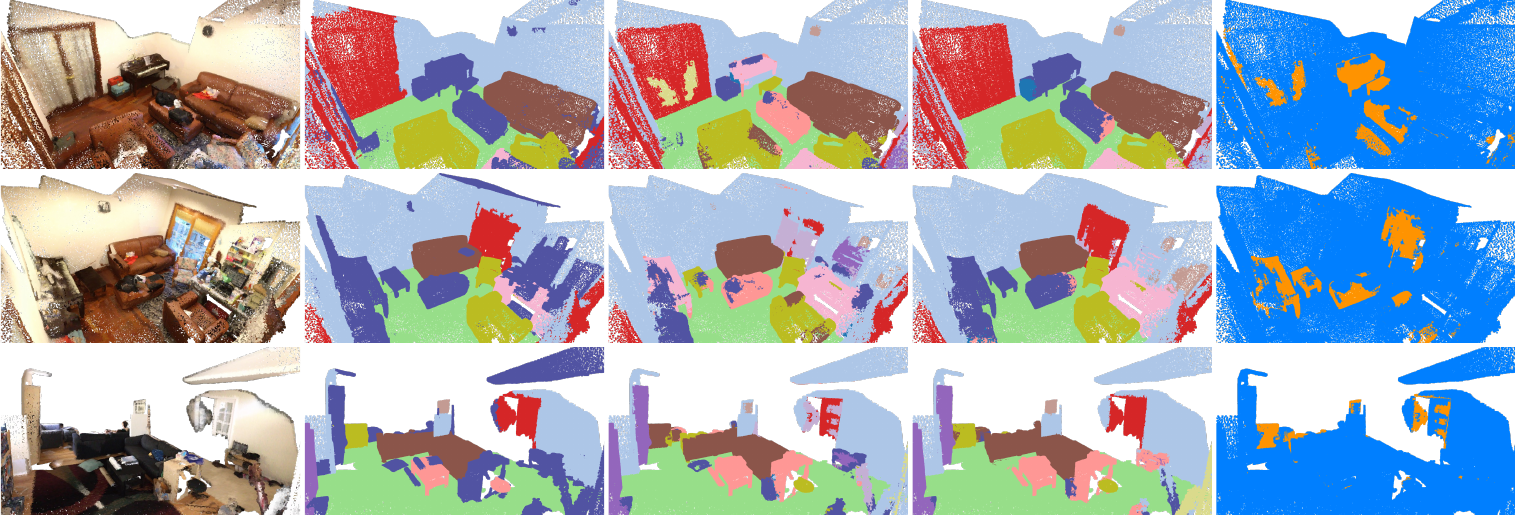}
\caption{Living rooms.}
\label{fig:demo-living}
\end{subfigure}

\begin{subfigure}{\linewidth}
    \includegraphics[width=\linewidth]{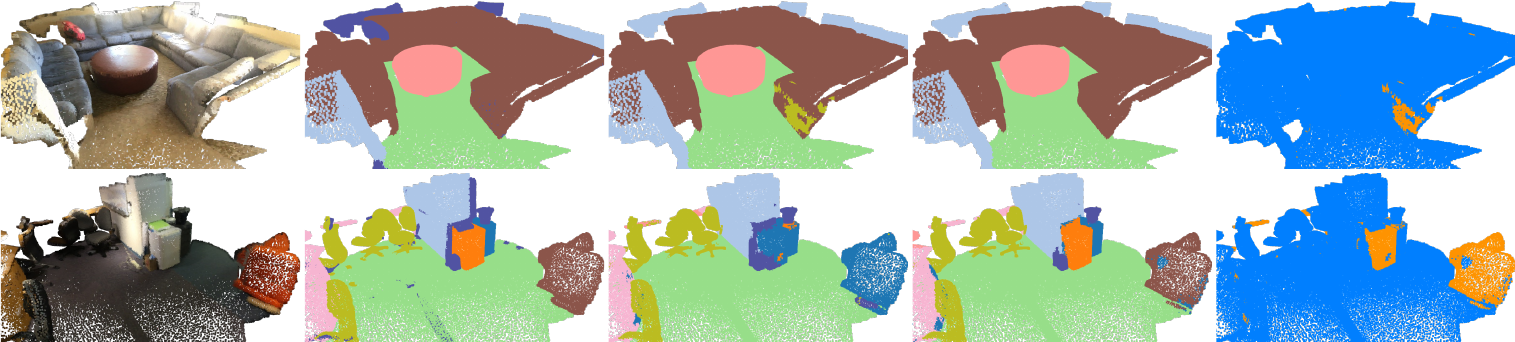}
\caption{Offices and lounges.}
\label{fig:demo-office}
\end{subfigure}

\caption{
We compare the results of baseline (lin.) with the proposed \method.
% The visualization is done on ScanNet.
}
\label{fig:demo-more}
\end{figure}

We provide additional qualitative results regarding the attention maps generated by our latent tokens in \cref{fig:attn-more}.
Although shared across stages, the latent tokens appear to attend to different parts of the scene, providing different scene context to the backbone models at different stages.
Consistently, the visual comparison with the baseline in \cref{fig:demo-more} shows clear improvements.
Across different scenes, \method yields cleaner and more coherent segmentations, with fewer confusions in cluttered regions and sharper boundaries at object interfaces.

\begin{table}[t]
\RawFloats
% \hfill
\resizebox{\linewidth}{!}{%
\begin{minipage}[t]{.47\textwidth}
\vspace{0pt}
\centering
%#################################################
% profiling
%#################################################
\resizebox{\linewidth}{!}{%
  \tablestyle{1pt}{1.05}
  \begin{tabular}{lrcc}
\toprule
%
% Compute Efficiency &\multicolumn{2}{c}{Params} &\multicolumn{3}{c}{ScanNet Val~\cite{scannet}} &\multicolumn{3}{c}{ScanNet200 Val~\cite{scannet200}} &\multicolumn{3}{c}{ScanNet++ Val~\cite{scannetpp}} &\multicolumn{3}{c}{S3DIS Area 5~\cite{s3dis}} &\multicolumn{3}{c}{S3DIS 6-fold~\cite{s3dis}} \\
% 
% \cmidrule(lr){1-1} \cmidrule(lr){2-3} \cmidrule(lr){4-6} \cmidrule(lr){7-9} \cmidrule(lr){10-12} \cmidrule(lr){13-15} \cmidrule(lr){16-18}
% 
Methods & Params. & Memory & Latency \\
\cmidrule(lr){1-4}
\multicolumn{4}{l}{\emph{Base model}} \\
\specialrule{0em}{1pt}{1pt}

% \rowstyle{\color{darkgray}}
Sonata (full.)~\cite{sonata}    & 124.8M    & 8.2G & 61.8ms \\

% \rowstyle{\color{darkgray}}
Sonata (ft.)                    & 108.5M    & 6.4G & 50.2ms \\

\cmidrule(lr){1-4}
\multicolumn{4}{l}{\emph{PEFT methods}} \\
\specialrule{0em}{1pt}{1pt}

Sonata (lin.)                           & 0.02M     & \best{6.4G} & \best{50.2ms} \\

% basic
% \ \ \plus BitFit~\cite{select_bitfit}   & 0.2M      & 6.4G & 50.2ms \\

\ \ \plus Adapter~\cite{adap}           & 2.8M      & \second 6.7G & \second 50.6ms \\

\ \ \plus LoRA~\cite{lora}              &  1.9M      & 7.9G & 52.8ms \\

\ \ \plus Prompt Tunning~\cite{prefix}  & 5.5M      & 7.3G & 51.5ms \\

\ \ \plus \best{\method (ours)}         & 1.8M      & 7.1G & 57.8ms \\

% - SA
% cuda_ms=40.123ms ???
% total alloc 67.242G

% - CA
% cuda_ms=60.501ms
% [overhead] Total alloc  mem: 96.399G
\bottomrule
\end{tabular}
}%
\caption{
  \textbf{Inference efficiency of PEFT methods.}
  We benchmark with the batch size fixed to 1.
}
  \label{tbl:profiling}
\end{minipage}
\hfill
% \hspace{.1em}
\begin{minipage}[t]{.49\textwidth}
\vspace{0pt}
\centering
%#################################################
% shape
%#################################################
\resizebox{\linewidth}{!}{%
  \tablestyle{1pt}{1.05}
  \begin{tabular}{lrcc}
\toprule
Shape-Part Seg. &\multicolumn{1}{c}{Params} &\multicolumn{2}{c}{ShapeNetPart~\cite{shapenetpart}}  \\
\cmidrule(lr){1-1} \cmidrule(lr){2-2} \cmidrule(lr){3-4}
Methods &\multicolumn{1}{c}{Learn.} & Cls. mIoU & Inst. mIoU \\

\cmidrule(lr){1-4}
\multicolumn{4}{l}{\emph{Full fine-tuning}} \\
\specialrule{0em}{1pt}{1pt}

ReCon (ft.)~\cite{shape_recon}          & 27.06             & 84.52 & 86.1  \\

\cmidrule(lr){1-4}
\multicolumn{4}{l}{\emph{PEFT methods}} \\
\specialrule{0em}{1pt}{1pt}

ReCon (lin.)~\cite{shape_recon}                 & 5.23M     & 83.06 & 85.2   \\

\ \ \plus PointLoRA~\cite{ptpeft_pointlora}     & 5.63M     & 83.98 & 85.4  \\
\ \ \plus PointGST~\cite{ptpeft_gst}       & 5.59M     & \second{83.98} & \second{85.8}  \\

\ \ \plus \best{\method (ours)}                 & 5.58M     & \best{84.02} & \best{85.8}  \\

\bottomrule
\end{tabular}
}%
\vspace{0pt}
% \vspace{14pt}
\caption{
  \textbf{Generalizing to 3D shapes.}
}
\label{tbl:partseg_shapenetpart}
\end{minipage}
% \hfill
}% resizebox
\end{table}
%##################################################################################################

%##################################################################################################
\begin{table}[t]
\RawFloats
% \hfill
\resizebox{\linewidth}{!}{%
\begin{minipage}[t]{.49\textwidth}
\vspace{0pt}
\centering
%#################################################
% conv
%#################################################
\resizebox{\linewidth}{!}{%
  \tablestyle{1pt}{1.05}
  \begin{tabular}{lrrccc}
\toprule
Tight Budgets &\multicolumn{2}{c}{Params} &\multicolumn{3}{c}{ScanNet200 Val~\cite{scannet200}}  \\
\cmidrule(lr){1-1} \cmidrule(lr){2-3} \cmidrule(lr){4-6}
Methods &\multicolumn{1}{c}{Learn.} &\multicolumn{1}{c}{Pct.} &mIoU &mAcc &allAcc \\

\cmidrule(lr){1-6}
\multicolumn{6}{l}{\emph{Training from scratch}} \\
\specialrule{0em}{1pt}{1pt}

SparseUnet~\cite{minkowski}         & 39.2M         & 100\%     & 25.0 & 32.9 & 80.4    \\

\cmidrule(lr){1-6}
\multicolumn{6}{l}{\emph{Full fine-tuning}} \\
\specialrule{0em}{1pt}{1pt}

% \rowstyle{\color{darkgray}}
SparseUnet (ft.)~\cite{msc}         & 0.02M         & 0.05\%    & 32.0 & 41.6 & 82.3    \\

\cmidrule(lr){1-6}
\multicolumn{4}{l}{\emph{PEFT methods with rank=1}} \\
\specialrule{0em}{1pt}{1pt}

SparseUnet (lin.)                       & 0.02M     & 0.05\%    & 1.5   & 2.5   & 53.6      \\
\ \ \plus BitFit~\cite{select_bitfit}   & 0.03M     & 0.07\%    & 4.8   & 6.9   & 62.7      \\
\ \ \plus Adapter~\cite{adap}           & 0.6M      & 1.5\%     & 9.5   & 13.4  & 69.5      \\
\ \ \plus LoRA~\cite{lora}              & 0.8M      & 2.0\%     & \second{13.2}  & \second{17.6}  & \second{74.7}      \\
\ \ \plus \best{\method (ours)}         & 0.6M      & 1.5\%     & \best{15.2}  & \best{22.9}  & \best{75.6}      \\

\bottomrule
\end{tabular}
}%
\caption{
  \textbf{Generalizing to convolutional networks.}
}
  \label{tbl:semseg_conv}
\end{minipage}
\hfill
% \hspace{.1em}
\begin{minipage}[t]{.49\textwidth}
\vspace{0pt}
\centering
%#################################################
% tight budgets
%#################################################
\resizebox{\linewidth}{!}{%
  \tablestyle{1pt}{1.05}
  \begin{tabular}{lrrccc}
\toprule
Tight Budgets &\multicolumn{2}{c}{Params} &\multicolumn{3}{c}{ScanNet Val~\cite{scannet}}  \\
\cmidrule(lr){1-1} \cmidrule(lr){2-3} \cmidrule(lr){4-6}
Methods &\multicolumn{1}{c}{Learn.} &\multicolumn{1}{c}{Pct.} &mIoU &mAcc &allAcc \\

\cmidrule(lr){1-6}
\multicolumn{6}{l}{\emph{Training from scratch}} \\
\specialrule{0em}{1pt}{1pt}

PTv3~\color{defaultcolor}{\cite{ptv3}}  & 124.8M    & 100\%     & 77.6 & 85.0 & 92.0    \\

\cmidrule(lr){1-6}
\multicolumn{6}{l}{\emph{Full fine-tuning}} \\
\specialrule{0em}{1pt}{1pt}

% \rowstyle{\color{darkgray}}
Sonata (full.)~\cite{sonata}            & 124.8M    & 100\%     & 79.4 & 86.1 & 92.5    \\

% \rowstyle{\color{darkgray}}
Sonata (ft.)                            & 108.5M    & 100\%     & 78.3 & 85.9 & 92.3    \\

\cmidrule(lr){1-6}
\multicolumn{4}{l}{\emph{PEFT methods with rank=1}} \\
\specialrule{0em}{1pt}{1pt}

% Sonata (lin.)                           &&&&& \\
Sonata (lin.)                           & 0.02M     & 0.02\%    & 72.5 & 83.1 & 89.7    \\
% basic
\ \ \plus Adapter~\cite{adap}           & 0.05M     & 0.04\%    & \second 74.0 & \second 84.1 & \second 90.5    \\

\ \ \plus LoRA~\cite{lora}              & 0.05M     & 0.05\%    & 42.5 & 55.4 & 75.1    \\

\ \ \plus Prompt Tunning~\cite{prefix}  & 0.05M     & 0.05\%    & 72.9 & 83.5 & 90.0    \\

% improvement - 
% \ \ \plus MAM~\cite{adap_mam}           & 5.9M      & 5.2\%    &  &  & &  & & & & & & & & & & & \\

% \ \ \plus Adapter+          &           &           &  &  & &  & & & & & & & & & & & \\

\ \ \plus \best{\method (ours)}         & 0.07M      & 0.06\%     & \best{75.2} & \best{84.8} & \best{91.1}    \\

\cmidrule(lr){1-6}
\multicolumn{4}{l}{\emph{PEFT methods with 0.1\% params.}} \\
\specialrule{0em}{1pt}{1pt}

% Sonata (lin.)                           & 0.02M     & 0.02\%    &  &  &  \\
Sonata (lin.)                           &&&&& \\

% basic
\ \ \plus Adapter~\cite{adap}           & 0.1M      & 0.1\%     & \second 75.1 & \second 84.7 & \second 91.1 \\

\ \ \plus LoRA~\cite{lora}              & 0.1M      & 0.1\%     & 75.0 & 84.4 & 91.1 \\

\ \ \plus Prompt Tunning~\cite{prefix}  & 0.1M      & 0.1\%     & 73.5 & 83.9 & 90.3 \\

% improvement - 
% \ \ \plus MAM~\cite{adap_mam}           & 5.9M      & 5.2\%    &  &  & &  & & & & & & & & & & & \\

% \ \ \plus Adapter+          &           &           &  &  & &  & & & & & & & & & & & \\

\ \ \plus \best{\method (ours)}         & 0.1M      & 0.1\%     & \best{76.5} & \best{85.5} & \best{91.6} \\

\cmidrule(lr){1-6}
% \multicolumn{4}{l}{\emph{PEFT methods}} \\
\multicolumn{4}{l}{\emph{PEFT methods with 1\% params.}} \\
\specialrule{0em}{1pt}{1pt}

Sonata (lin.)                           &&&&& \\
% Sonata (lin.)                           & 0.02M     & 0.02\%    & 72.5 & 83.1 & 89.7    \\

% basic
\ \ \plus Adapter~\cite{adap}           & 1.1M      & 1.0\%     & 76.6 & 85.3 & 91.7    \\

\ \ \plus LoRA~\cite{lora}              & 1.1M      & 1.0\%     & \second 76.7 & \second 85.6 & \second 91.7    \\

\ \ \plus Prompt Tunning~\cite{prefix}  & 1.1M      & 1.0\%     & 73.8 & 84.2 & 90.4    \\

% improvement - 
% \ \ \plus MAM~\cite{adap_mam}           & 5.9M      & 5.2\%    &  &  & &  & & & & & & & & & & & \\

% \ \ \plus Adapter+          &           &           &  &  & &  & & & & & & & & & & & \\

\ \ \plus \best{\method (ours)}         & 1.1M      & 1.0\%     & \best{78.2} & \best{86.3} & \best{92.2}    \\

\bottomrule
\end{tabular}
}%
\vspace{0pt}
% \vspace{14pt}
\caption{
  \textbf{Performance with tight budgets.}
  % Method with $^*$ denotes our re-produced results.
}
\label{tbl:semseg_tight}
\end{minipage}
% \hfill
}% resizebox
\end{table}
%##################################################################################################

% }%
%%%%%%%%%%%%%%%%%%%%%%%%%%%%%%%%%%%%%%%%%%%%%%%%%%%%%%%%%%%%%%%%%%%%%%%%%%%%%%%%%%%%%%%%%%%%%%%%%%%%%%%%%%%%%%%%%%%%%%%%

% {\color{white}.}

{\small
\bibliographystyle{ieee_fullname_order}
\bibliography{ref}
}

\end{document}